\documentclass[review]{elsarticle}

\usepackage{lineno,hyperref,xcolor,todonotes}
\modulolinenumbers[5]

\journal{Neurocomputing}

\usepackage{float}
\usepackage{graphicx}
\usepackage{caption}
\usepackage{subcaption}
\usepackage{amsmath}
\usepackage{amssymb}
\usepackage{booktabs} 
\usepackage{hyperref}
\usepackage[utf8x]{inputenc}
\definecolor{darkgreen}{rgb}{0.0, 0.5, 0.0}
\begin{document}

\begin{frontmatter}

\title{Calibration of Deep Probabilistic Models with Decoupled Bayesian Neural Networks}

\author[add1]{Juan Maro\~nas \corref{mycorrespondingauthor}}
\author[add1]{Roberto Paredes \fnref{myfootnote}}
\author[add2]{Daniel Ramos \fnref{myfootnote}}

\address[add1]{PRHLT - Pattern Recognition and Human Language Technology Research Center, Universitat Politècnica de Valencia, Spain}
\address[add2]{AUDIAS - Audio Data Intelligence and Speech, Universidad Autónoma de Madrid, Spain}

\cortext[mycorrespondingauthor]{Corresponding author: $<$jmaronas@prhlt.upv.es$>$ $<$jmaronasm@gmail.com$>$ . Pattern Recognition and Human Language Technologies Research Center, Universitat Politècnica de València, Camino de Vera, s/n, Valencia 46022, Spain }
\fntext[myfootnote]{Equal contribution.  Alphabetical order.}

\begin{abstract}
Deep Neural Networks (DNNs) have achieved state-of-the-art accuracy performance in  many  tasks. However, recent works have pointed out that the outputs provided by these models are not well-calibrated, seriously limiting their use in critical decision scenarios. In this work, we propose to use a decoupled Bayesian stage, implemented with a Bayesian Neural Network (BNN), to map the uncalibrated probabilities provided by a DNN to calibrated ones, consistently improving calibration. Our results evidence that incorporating uncertainty provides more reliable probabilistic models, a critical condition for achieving good calibration. We report a generous collection of experimental results using high-accuracy DNNs in standardized image classification benchmarks, showing the good performance, flexibility and robust behaviour of our approach with respect to several state-of-the-art calibration methods. Code for reproducibility is provided.
\end{abstract}


\end{frontmatter}


\section{Introduction}
Deep Neural Networks (DNNs) represent the state-of-the-art performance in many tasks such as image classification  \citep{DBLP:journals/corr/HuangLW16a,DBLP:journals/corr/ZagoruykoK16}, language modeling \citep{DBLP:journals/corr/abs-1301-3781,DBLP:journals/corr/MikolovSCCD13}, machine translation \citep{DBLP:journals/corr/VaswaniSPUJGKP17} or speech recognition \citep{hinton16speechprocessing}. As a consequence, DNNs are nowadays used as important parts of complex and critical decision systems.

However, although accuracy is a suitable measure of the performance of DNNs in numerous scenarios, there are many applications in which the probabilities provided by a DNN must be also \emph{reliable}, i.e. well-calibrated \cite{dawid82wellCalibratedBayesian}. This is mainly because well-calibrated DNN output probabilities present two important and interrelated properties: First, they can be \emph{reliably} interpreted as probabilities \cite{dawid82wellCalibratedBayesian} enabling its adequate use in Bayesian decision making. Second, calibrated probabilities lead to \emph{optimal} expected costs in any Bayesian decision scenario, regardless of the choice of the costs of wrong decisions \cite{cohen04calibrated,brummer10PhD}. 

As an example, if we assist a critical decision process, e.g. a medical diagnosis pipeline where a human practitioner uses the information of a machine learning model, the human needs that the probabilities provided by the model are interpretable \citep{Caruana:2015:IMH:2783258.2788613}. In such cases,  supporting the decision of an expert practitioner with an uncalibrated probability  (e.g. $0.9$ probability that a medical image does not present a malign brain tumor) can have drastic consequences as our model will not be reflecting the true proportion of real outcomes. 

Apart from the medical field, see \cite{Caruana:2015:IMH:2783258.2788613} for details, many other applications can benefit from well-calibrated probabilities, which has motivated the machine learning community towards exploring different techniques to improve calibration performance in different contexts \citep{Caruana:2015:IMH:2783258.2788613,zadrozny02,niculeskuMizil05predictingGoodProbabilities}. For instance, applications where predictions consider different probabilistic models that must be combined, such as neural networks and language models for machine translation \citep{Gulcehre:2017:ILM:3103639.3103741}; applications with a big mismatch between training and test distributions, as in speaker and language recognition \citep{brummer10PhD,brummer06calibrationLanguage}; self-driving cars \cite{journals/corr/BojarskiTDFFGJM16}; out-of-distribution sample detection \cite{Lee2017TrainingCC}; and so on. 

One classical way of improving calibration is by optimizing an expected value of a proper scoring rule (PSR) \cite{niculeskuMizil05predictingGoodProbabilities,deGroot83forecasters,NIPS2017_7219}, such as the logarithmic scoring rule (whose average value is the cross-entropy or negative log-likelihood, NLL) and the Brier scoring rule (whose average value is an estimate of the mean squared error). However, a proper scoring rule not only measures calibration, but also the ability of a classifier to discriminate between different classes, a magnitude known as \emph{discrimination} or \emph{refinement} \cite{deGroot83forecasters,brummer10PhD,ramos18crossEntropy}, which is necessary to achieve good accuracy values \cite{brummer10PhD}. Both quantities are indeed additive up to the value of the average PSR. Thus, optimizing the average PSR is not a guarantee of improving calibration, because the optimization process could lead to worse calibration at the benefit of an improved refinement. This effect has been recently pointed-out in DNNs \cite{DBLP:journals/corr/GuoPSW17}, where models trained to optimize the NNL have outstanding accuracy but are bad calibrated towards the direction of over-confident probabilities. Here, over-confidence means that, for instance, all samples of a given class where the confidence given by the DNN was around $0.99$, are correctly classified in much less than $99\%$ of the cases.

Motivated by this observation, several techniques have been recently proposed to improve the calibration of DNNs while aiming at preserving their accuracy  \cite{NIPS2017_7219,DBLP:journals/corr/GuoPSW17,DBLP:conf/icml/KuleshovFE18,pmlr-v80-kumar18a,1809.10877}, basing their design choice on point estimate approaches, e.g maximum likelihood. However,  as we will justify in the next section, a proper address of uncertainty, as done by Bayesian approaches, is a clear advantage towards reliable probabilistic modelling; a fact that has been recently shown for example in the context of computer vision \cite{NIPS2017_7141}. Despite these well-known properties of Bayesian statistics, they have received major criticisms when they are used in DNN pipelines, mainly due to important limitations such as prior selection, memory and computational costs, and inaccurate approximations to the distributions involved \cite{NIPS2017_7219,DBLP:conf/icml/KuleshovFE18,pmlr-v80-kumar18a,fixing}.

\begin{figure}[!t]
    \centering
    \begin{subfigure}[t]{\linewidth}
    \includegraphics[width=\textwidth]{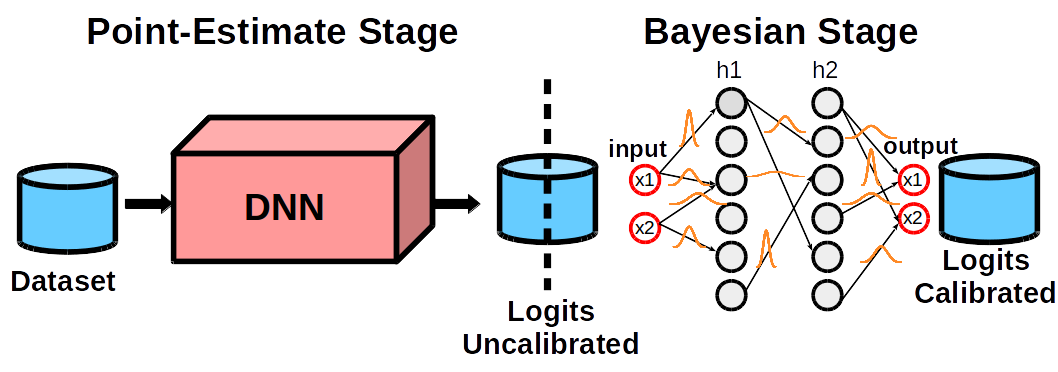}
    \caption{An example of the architecture of our proposed model. On the left top figure, an expensive DNN is trained on a dataset. Then, the (uncalibrated) output of such DNN is the input to the BNN calibration stage. The inputs and outputs of the Bayesian stage have the same dimensionality (given by the number of classes). Orange Gaussians on each arrow represent the variational distributions on each parameter.}
    \end{subfigure}
    \begin{subfigure}[t]{\linewidth}
        \includegraphics[width=\textwidth]{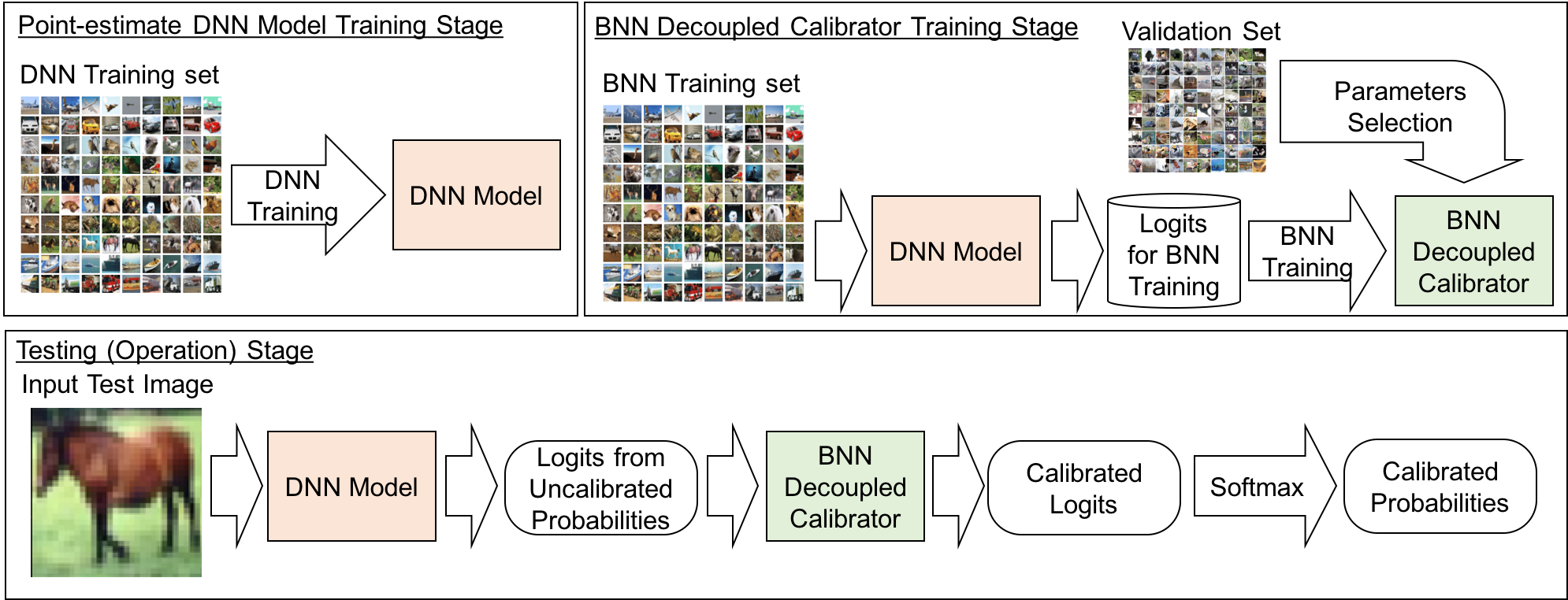}
        \caption{This figure represents a description of the training, validation and test stages of the proposed model.}
    \end{subfigure}
    \caption{A graphical description of the proposed architecture}
    \label{fig:proposed_architechture}
\end{figure}

In this work we aim at bridging this gap, i.e. being able to combine the state-of-the-art accuracy performance provided by DNNs, with the good properties of Bayesian approaches towards principled probabilistic modelling. Following this objective, we propose a new procedure to use Bayesian statistics in DNN pipelines, without compromising the whole system performance. The main idea is to re-calibrate the outputs (in the form of logits) of a pre-trained DNN, using a decoupled Bayesian stage which we implement with a Bayesian Neural Network (BNN), as shown in figure \ref{fig:proposed_architechture}. 

\begin{figure}[!t]
    \centering
 \begin{subfigure}[l]{0.32\columnwidth}
 \subcaption*{Uncalibrated}
  \includegraphics[width=1.0\columnwidth]{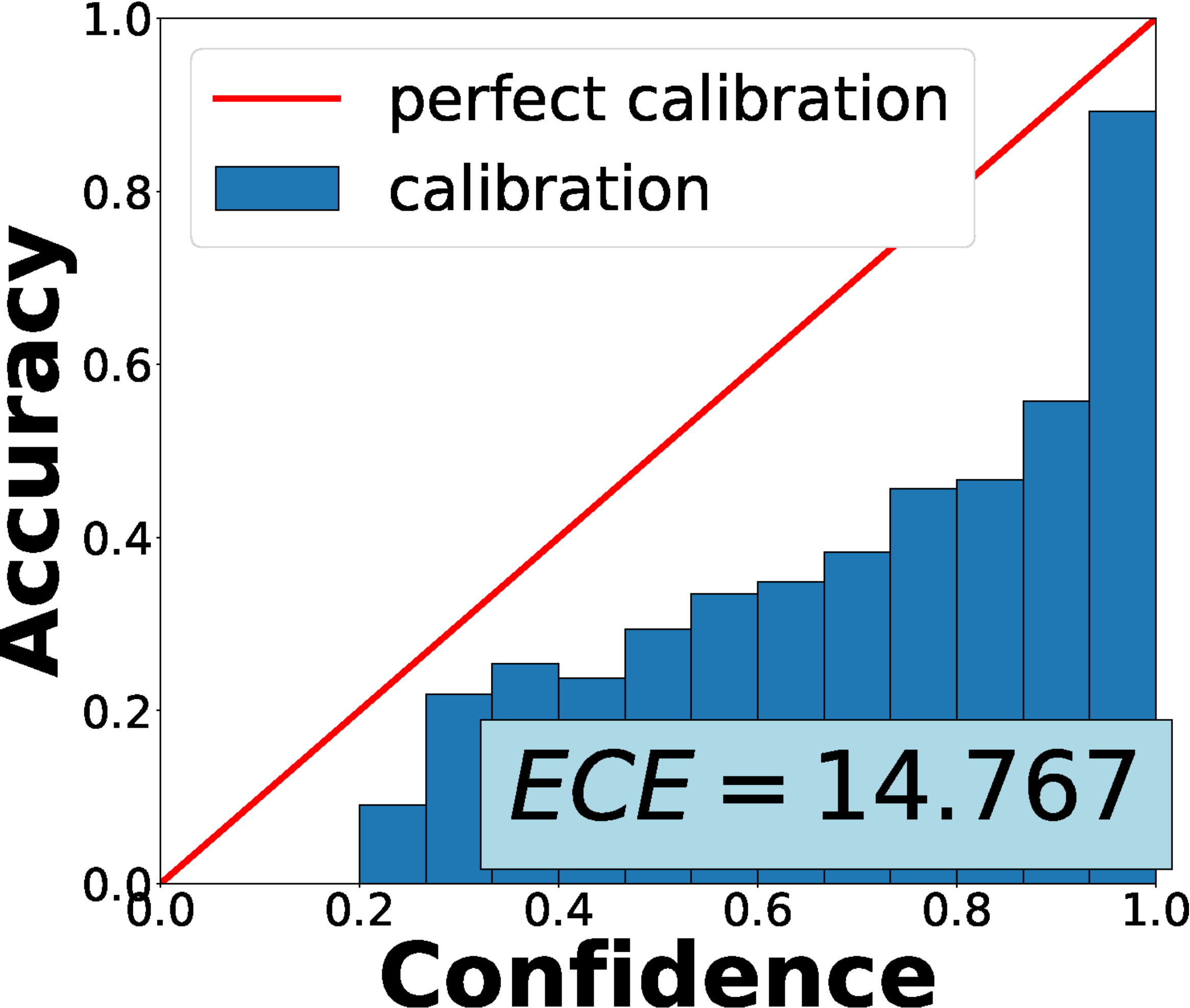}
\end{subfigure}
\begin{subfigure}[c]{0.32\columnwidth}
\subcaption*{Temperature-Scaling}
  \includegraphics[width=1.0\columnwidth]{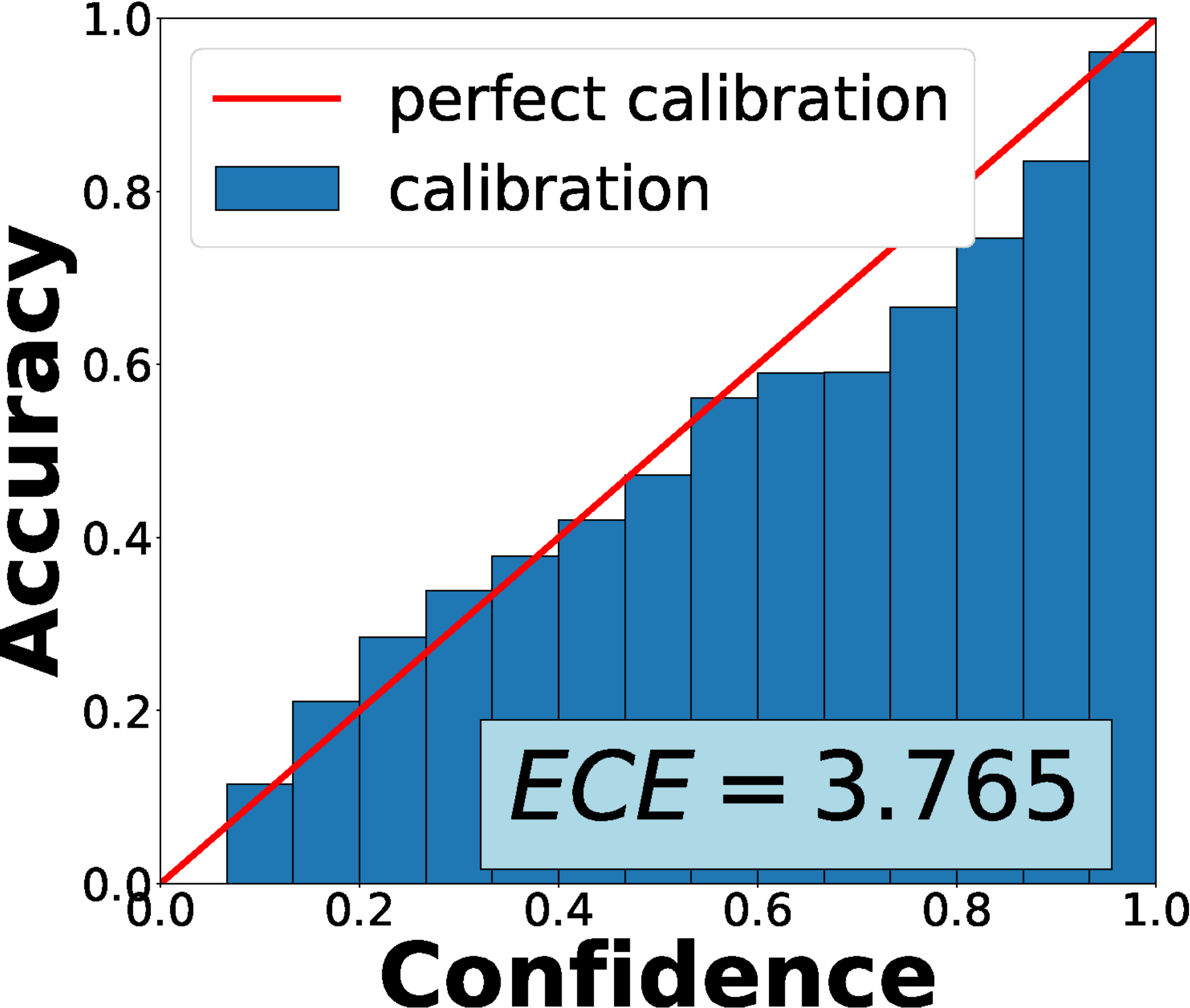}
\end{subfigure}
\begin{subfigure}[r]{0.32\columnwidth}
\subcaption*{Bayesian Neural Network}
  \includegraphics[width=1.0\columnwidth]{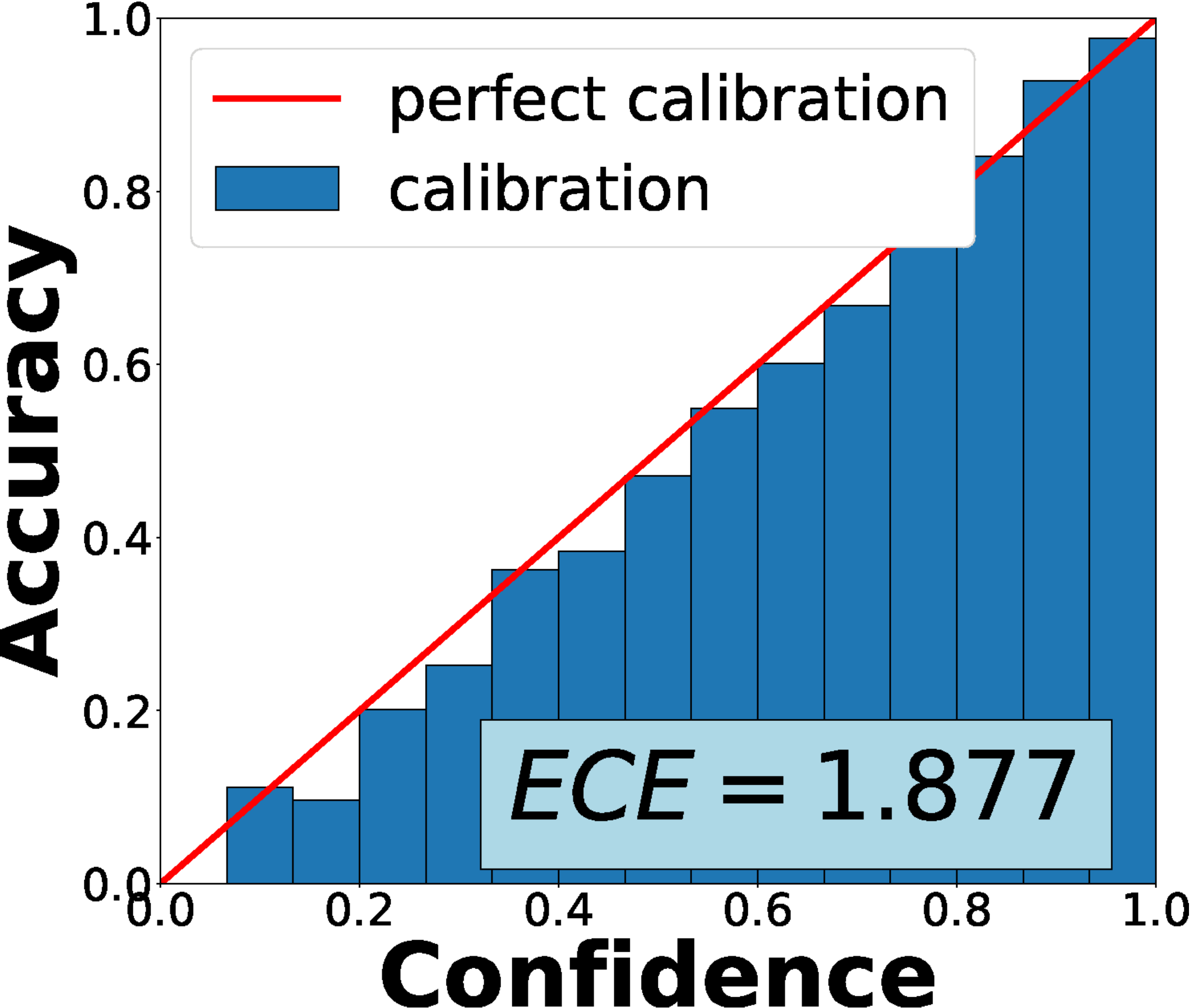}
\end{subfigure}\\
\begin{subfigure}[l]{0.32\columnwidth}
  \includegraphics[width=1.0\columnwidth]{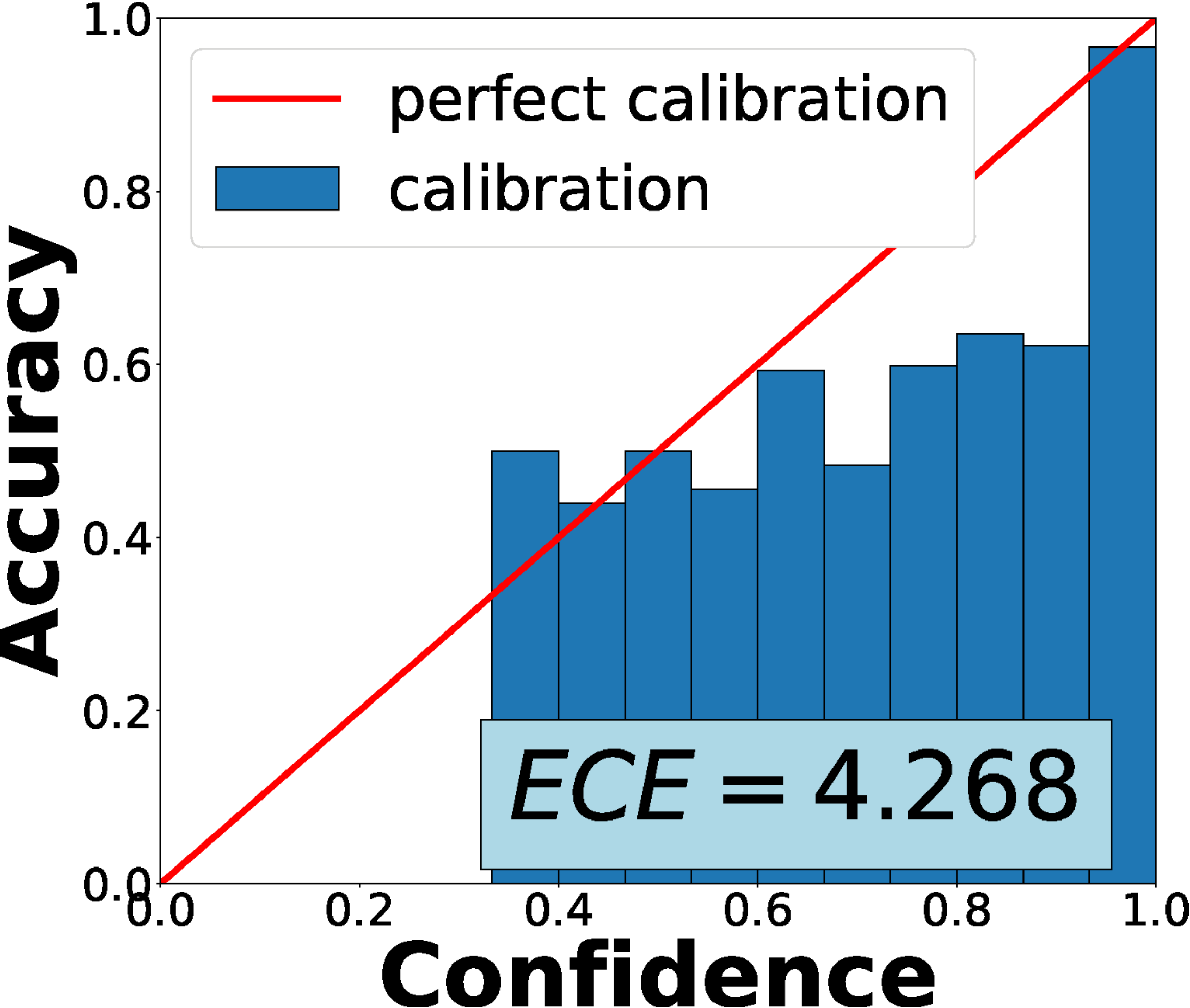}\end{subfigure}
\begin{subfigure}[c]{0.32\columnwidth}
  \includegraphics[width=1.0\columnwidth]{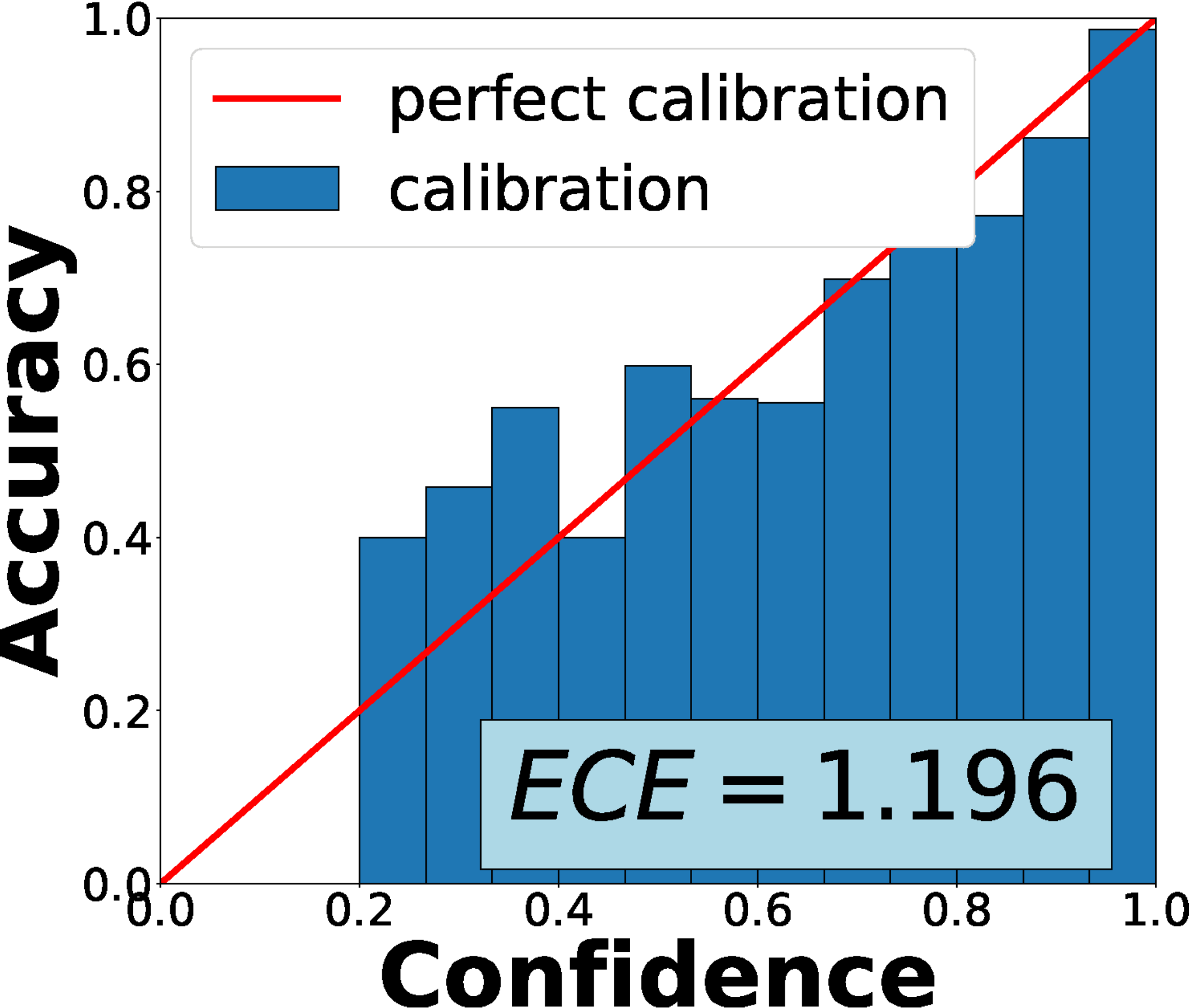}
\end{subfigure}
\begin{subfigure}[r]{0.32\columnwidth}
  \includegraphics[width=1.0\columnwidth]{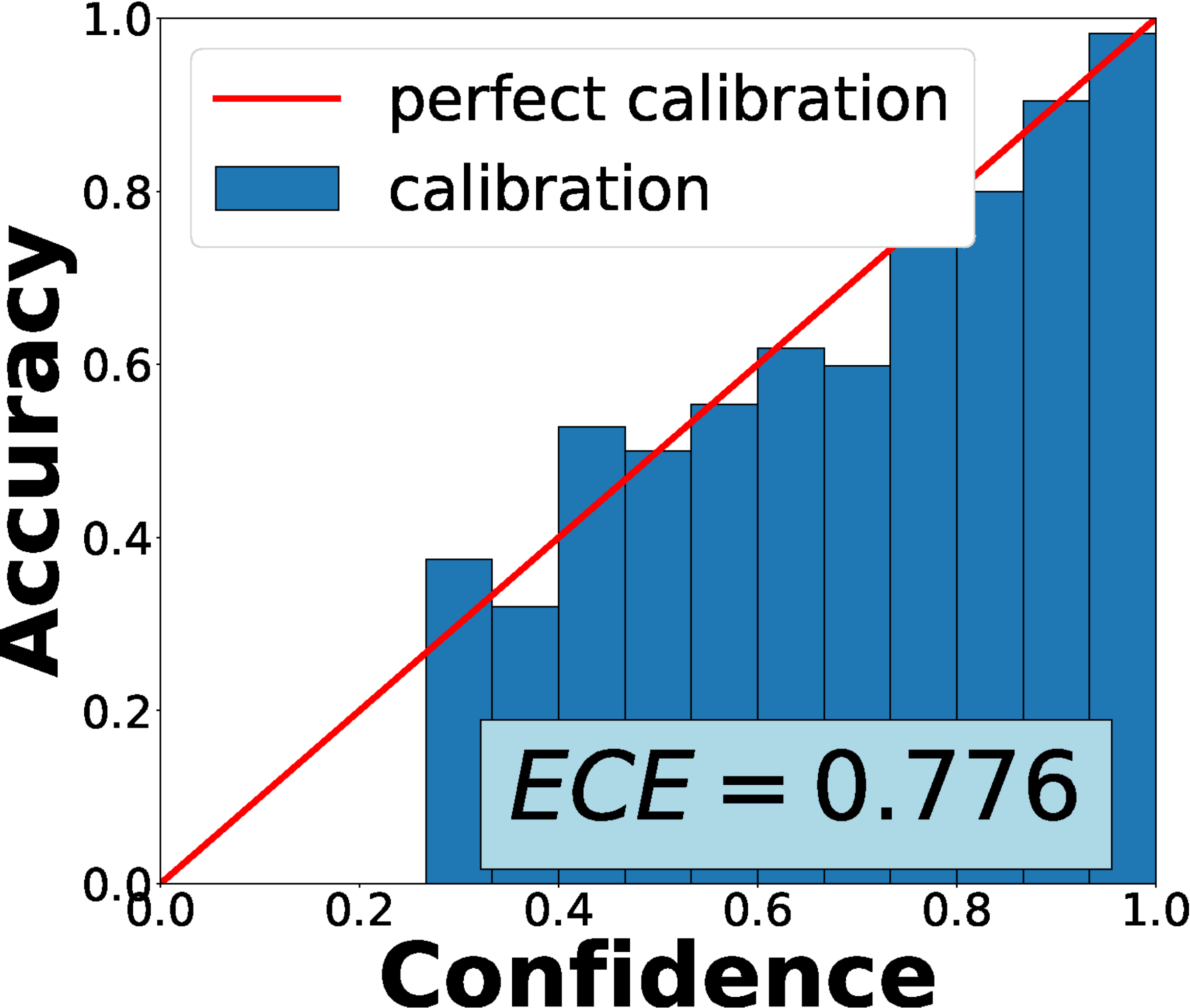}
\end{subfigure}
\caption{Reliability diagrams \cite{DBLP:journals/corr/GuoPSW17} for two DNNs trained on two computer vision benchmarks, namely CIFAR-100 (top row) and CIFAR-10 (bottom row). Column titles indicate the calibration technique. The red $x=y$ line represents perfect calibration. The closer the histogram to the line, the better the calibration of the technique. We complement the plot with the Expected Calibration Error (ECE \%) for 15 bins. The lower the ECE value, the better the calibration of the technique. See experimental section for a more detailed description of this performance measure.}
\label{fig:reliability_diagram}
\end{figure}

This approach presents clear advantages, including: better performance than other state-of-the-art calibration techniques for DNNs, such as Temperature Scaling (TS) \cite{DBLP:journals/corr/GuoPSW17}(see figure \ref{fig:reliability_diagram}); scalability with the data size and the complexity of the pre-trained DNN both during training and test phases, as BNNs can be trained to re-calibrate any pre-trained DNN regardless of its architecture or type; and robustness, since the approach works consistently well in a numerous variety of experimental set-ups and training hyperparameters. One important conclusion drawn from this work is that as long as the uncertainty is properly addressed, we can improve the calibration performance making use of complex models. This observation contrasts with the main argument from \cite{DBLP:journals/corr/GuoPSW17}, where the authors argue that TS, their best-performing method, worked better than complex models because the calibration space is inherently simple, and complex models tend to over-fit. It should be noted that this observation can be wrong in its origin, as the calibration space can be application-dependent, which motivates the necessity of developing complex models that can perform in different scenarios. 

The work is organized as follows. We begin by introducing and motivating the Bayesian framework for reliable probabilistic modelling in the classification scenario. We then describe the steps involved in the BNN-based approach considered in this work. We finally report a wide set of experiments to support our hypotheses.

\section{Related Work}

From a list of classical methods to improve calibration (such as Histogram Binning \citep{Zadrozny:2001:OCP:645530.655658}, Isotonic Regression \citep{zadrozny02}, Platt Scaling \citep{Platt99probabilisticoutputs}, Bayesian Binning into Quantiles \citep{Naeini:2015:OWC:2888116.2888120}); TS \citep{DBLP:journals/corr/GuoPSW17} has been reported as one of the best techniques for the computer vision tasks of interest in our current work. On the other hand, there are several works that study overconfident predictions and model uncertainty in different contexts, but without reporting an explicit measurement of calibration performance in DNNs. For instance, \cite{mcdropoutgal} link Gaussian processes with classical dropout regularized networks, showing how uncertainty estimates can be obtained from these networks. Indeed, the authors themselves state that these Bayesian outputs are not calibrated. In \cite{Pereyra2017RegularizingNN}, an entropy term is added to the log-likelihood to relax overconfidence. \cite{NIPS2017_7219} propose training network ensembles with adversarial noise samples to output confident scores. In \cite{DBLP:journals/corr/abs-1805-05396}, a confidence score is obtained by using the probes of the individual layers of the neural network classifier. In \cite{DeVries2018LearningCF}, the authors propose to train a second confident output, obtained from the penultimate layer of the classifier, by interpolation of the softmax output and the true value, scaled by this score. \cite{Lee2017TrainingCC} propose a generative approach for detecting out-of-distribution samples but evaluate calibration performance comparing their method with TS as the decoupled calibration technique.

On the side of BNNs,  \cite{DBLP:journals/corr/GalG15a} connect Bernoulli dropout with BNNs, and  \cite{NIPS2015_5666} formalize Gaussian dropout as a Bayesian approach. In \cite{1703.01961}, novel BNNs are proposed, using RealNVP \cite{45819} to implement a normalizing flow \cite{1505.05770}, auxiliary variables \citep{Maaloe:2016:ADG:3045390.3045543} and local reparameterization \citep{NIPS2015_5666}. None of these approaches measure calibration performance explicitly on DNNs, as we do. For instance, \cite{1703.01961} and \cite{NIPS2017_7219} evaluate uncertainty by training on one dataset and use it on another, expecting a maximum entropy output distribution. More recently, \cite{DBLP:journals/corr/abs-1805-10377} propose a scalable inference algorithm that is also asymptotically accurate as MCMC algorithms and \cite{fixing} propose a deterministic way of computing the ELBO to reduce the variance of the estimator to 0, allowing for faster convergence. They also propose a hierarchical prior on the parameters. 

\section{Bayesian Modelling and Calibration}
\label{BayesianModellingandCalibration}
We start by describing calibration in a class-conditional classification scenario as the one explored in this work and highlighting the importance of using Bayesian modelling. This will allow us to motivate our proposed framework, introduced in the next section. Although we focus on class-conditional modelling, many of the claims covered in this section apply to any probability distribution we wish to assign from data. 

In a classification scenario, calibration can be intuitively described as the agreement between the class probabilities assigned by a model to a set of samples, and the proportion of those classified samples where that class is actually the true one. In other words, if a model assigns a class $t$,  with probability $0.8$ to each sample $x$ in a set of samples, we expect that $80\%$ of these samples actually belong to class $t$ \cite{dawid82wellCalibratedBayesian,zadrozny02}. In addition, we require our probability distributions to be sharpened, meaning that the probability mass is concentrated only in some of the classes (ideally only in the correct class for each sample). This allows the classifier to separate the different classes efficiently. It should be noted that a classifier that presents bad discrimination can be useless even if it is perfectly calibrated, for instance, a prior classifier. On the other hand, uncertainty quantification (for instance for out-of-distribution-samples (ood) or for input-corrupted-samples detection) has strong relations with calibrated distributions. Note that for a set of ood samples evaluated over a $C$-class problem, where on average we have $\frac{1}{C}$ accuracy, a calibrated model will assign probability $\frac{1}{C}$. Thus, the average entropy would be the maximum entropy, and thus uncertainty about this input would be maximal, as expected from a good uncertainty quantifier. 

Formally, our objective is to assign a probability distribution  $\hat{p}(t|x)$  having observed a set  $\mathcal{O}=\{(x_i,t_i)\}_{i=1}^N$ of training samples, where $i$ denotes the training sample index. With this model, we then assign a categorical label $t^*$  to a test sample $x^*$, a decision made taking into account the probability distribution of the different class labels given the sample. For simplicity we assign the label $t^*$ to the most probable category\footnote{We adopt this maximum-a-posteriori (MAP) decision scheme for simplicity although, in a strict Bayesian decision scenario, MAP assumes equal losses for each wrong class decision, and prior probabilities equal to the empirical proportions of each class in the training data. In scenarios where classes have different importance or the empirical proportions of training and testing datasets differ, this MAP decision rule can be wrong in origin.}. The value of $\hat{p}(t^*|x^*)$ for the selected class is also referred to as the \emph{confidence} on the decision of the classifier.

Our main objective is providing a model $\hat{p}(t|x)$  that is most consistent with the data distribution $p(t|x)$ as it is well known that the lower the gap between $\hat{p}(t|x)$ and $p(t|x)$, the closer we are to an optimal Bayesian decision rule. This better representation of $p(t|x)$ will be reflected as better probability estimates and thus better calibration properties; and can be achieved by incorporating parameter uncertainty in the predictions, which is the difference between Bayesian and point-estimate models.

We denote $\theta$ as the model parameters vector from a parameter space $\Theta$, e.g. the weights of a neural network. A point-estimate approach assigns $\hat{p}(t|x)$ by selecting the value $\hat{\theta}$ that optimizes a criterion given the observations $\mathcal{O}$. Thus, the probability is assigned through:
\begin{equation}
\begin{split}
    \hat{\theta}=\underset{\theta \in \Theta}{\mbox{argmax}}\,\, L(\theta,\mathcal{O})\\
    \hat{p}(t|x)=p(t|x,\hat{\theta})
\end{split}
\end{equation}

Here, $L(\theta,\mathcal{O})$ is the maximum likelihood (ML) or the maximum a posterior (MAP) distributions. For MAP optimization we have: 
\begin{equation}
    L(\theta,\mathcal{O})= \frac{1}{N}\overset{N}{\underset{i}{\sum}}\mbox{CE}(x_i,t_i,\theta) + \log p(\theta)\label{equation_MAP},
\end{equation}
where for ML the $\log p(\theta)$ is removed from the loss function. $\mbox{CE}$ denotes the cross-entropy function, which is derived from the assumption of a categorical likelihood  i.e. $t\sim \mbox{Cat}(t|x)$. As a consequence, the prediction is entirely based on a particular choice of the value of the parameter vector $\theta$, even though the loss function can have several different local minima in different values in $\Theta$.

On the other hand, in a Bayesian paradigm, predictions are done by marginalizing all the model parameters:

\begin{equation}
\centering
\hat{p}(t|x) =p(t|x,\mathcal{O})=\mathbb{E}_{p(\theta|\mathcal{O})}[p(t|x,\theta)],
\label{bayesian_integral}
\end{equation}

\noindent which is no more than the expected value of all the likelihood models $p(t|x,\theta)$ under the posterior distribution $p(\theta|\mathcal{O})$ of the parameters given the observations:

\begin{equation}
    p(\theta|\mathcal{O})=\frac{\underset{i}{\prod} p(t_i|x_i,\theta) \cdot p(\theta)}{\int_\Theta \,d\theta\,\, \underset{i}{\prod} p(t_i|x_i,\theta) \cdot p(\theta)}
    \label{bayesian_posterior}
\end{equation}

Here, we assume that the input distribution $p(x|\theta)$ is not modelled. From both equations \ref{bayesian_integral} and \ref{bayesian_posterior}, it is clear that the Bayesian model incorporates parameter uncertainty, given by the posterior distribution, through a weighted average of the different likelihoods in equation \ref{bayesian_integral}. The importance given to each likelihood is directly related to its consistency with the observations (as given by the likelihood term in the numerator from equation \ref{bayesian_posterior})\footnote{This claim can be done by considering a non-informative prior $p(\theta)$, which we do here for simplicity.}.

Considering just Bayesian class-conditional models and keeping in mind the expressions involved in computing the posterior, we should expect the following behaviour: models that are likely to represent a region of the input space where only samples from a particular class are present will end up assigning high confidence to that particular class in that region, because increasing the density towards other classes will not raise the likelihood from the numerator in equation \ref{bayesian_posterior}. On the other hand, models that are likely to explain regions where features from two or more classes overlap will be forced to increase the probability density of both classes, thus \emph{relaxing} the ultimate confidence provided to those classes in that region of the input space. This behaviour will favour probabilities that closely reflect the patterns showed in the data, and thus we will be achieving our ultimate goal discussed at the beginning of this section. Moreover, note that apart from providing more accurate confidence values, Bayesian models will also consider underrepresented parts of the input space, as given by the corresponding amount of density placed by the posterior on the set of parameters that explain these regions. By definition, point estimate approaches will not present any of these mentioned effects.

\begin{figure}[!t]
    \centering
 \begin{subfigure}[l]{0.35\columnwidth}
  \includegraphics[width=1.0\columnwidth]{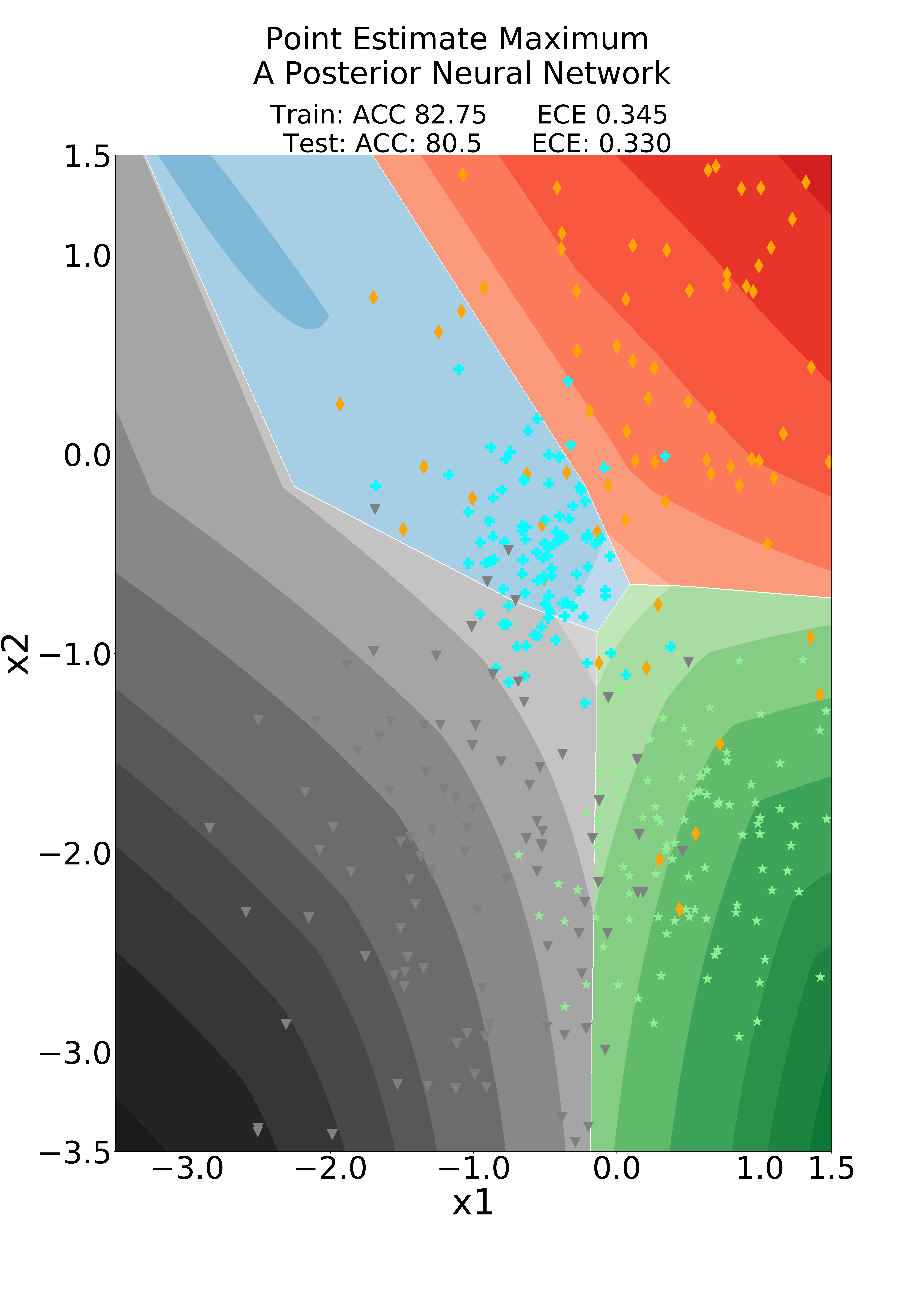}
\end{subfigure}
\begin{subfigure}[r]{0.35\columnwidth}
  \includegraphics[width=1.0\columnwidth]{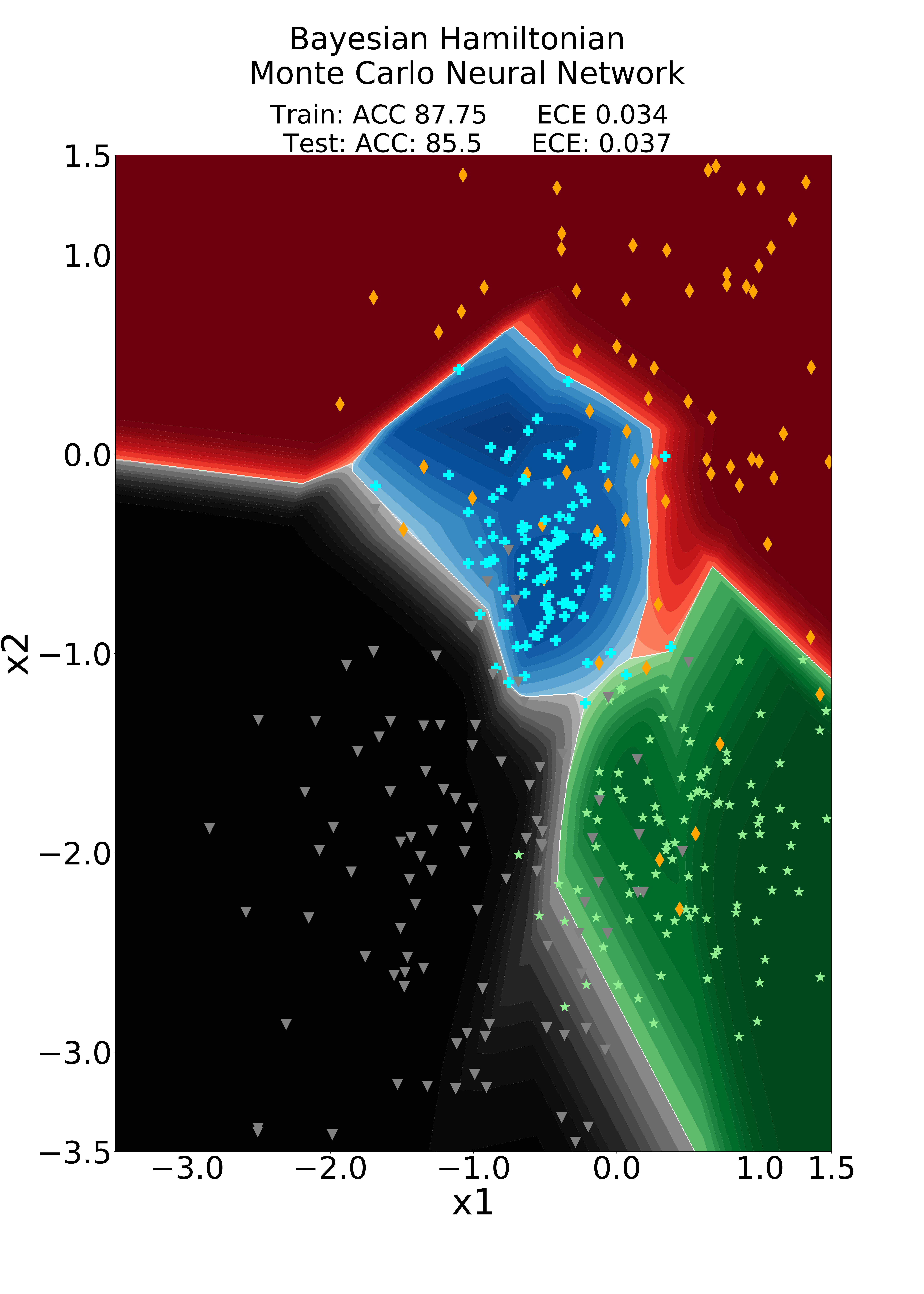}
\end{subfigure}\\
\begin{subfigure}[l]{0.35\columnwidth}
  \includegraphics[width=1.0\columnwidth]{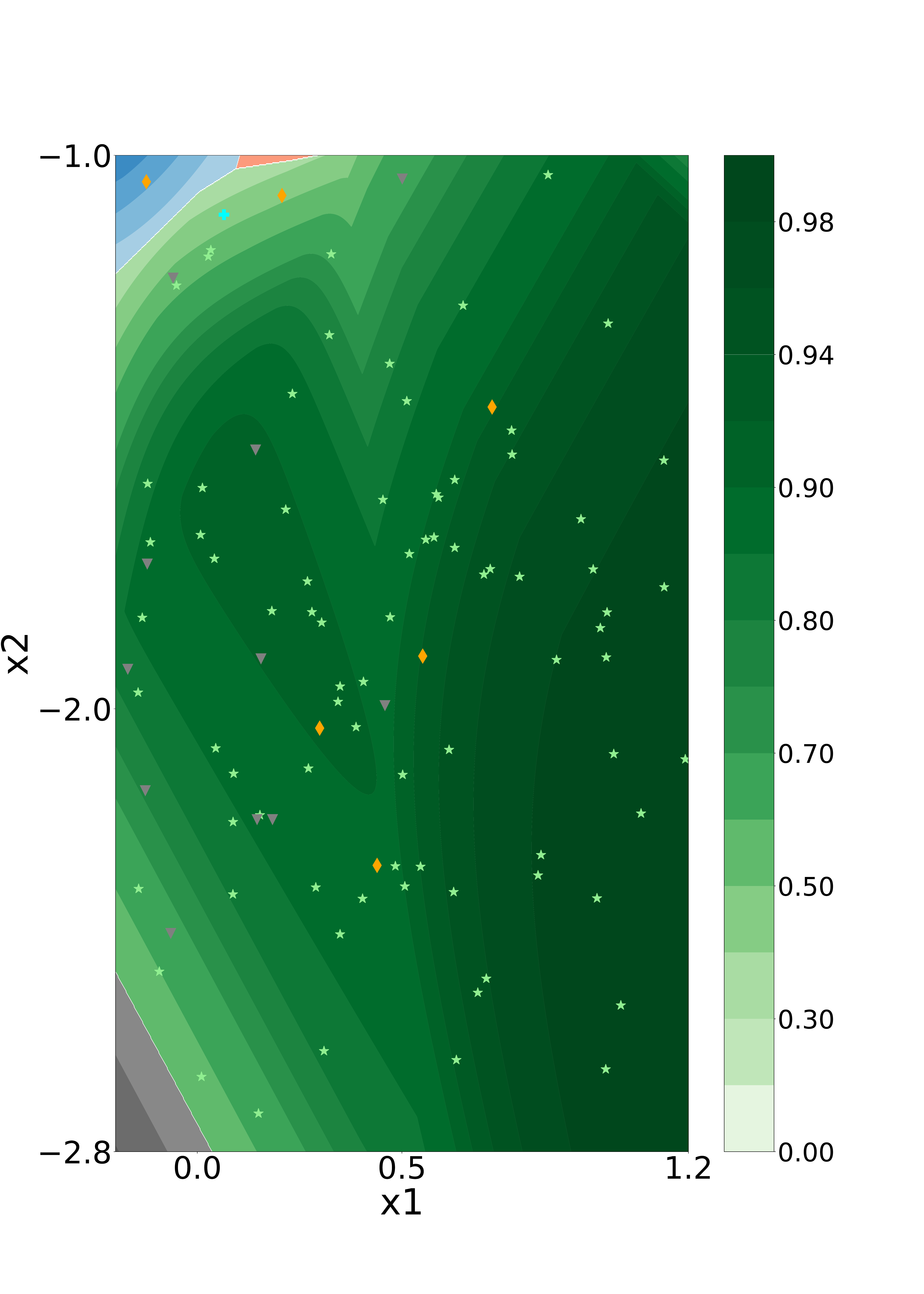}
\end{subfigure}
\begin{subfigure}[r]{0.35\columnwidth}
  \includegraphics[width=1.0\columnwidth]{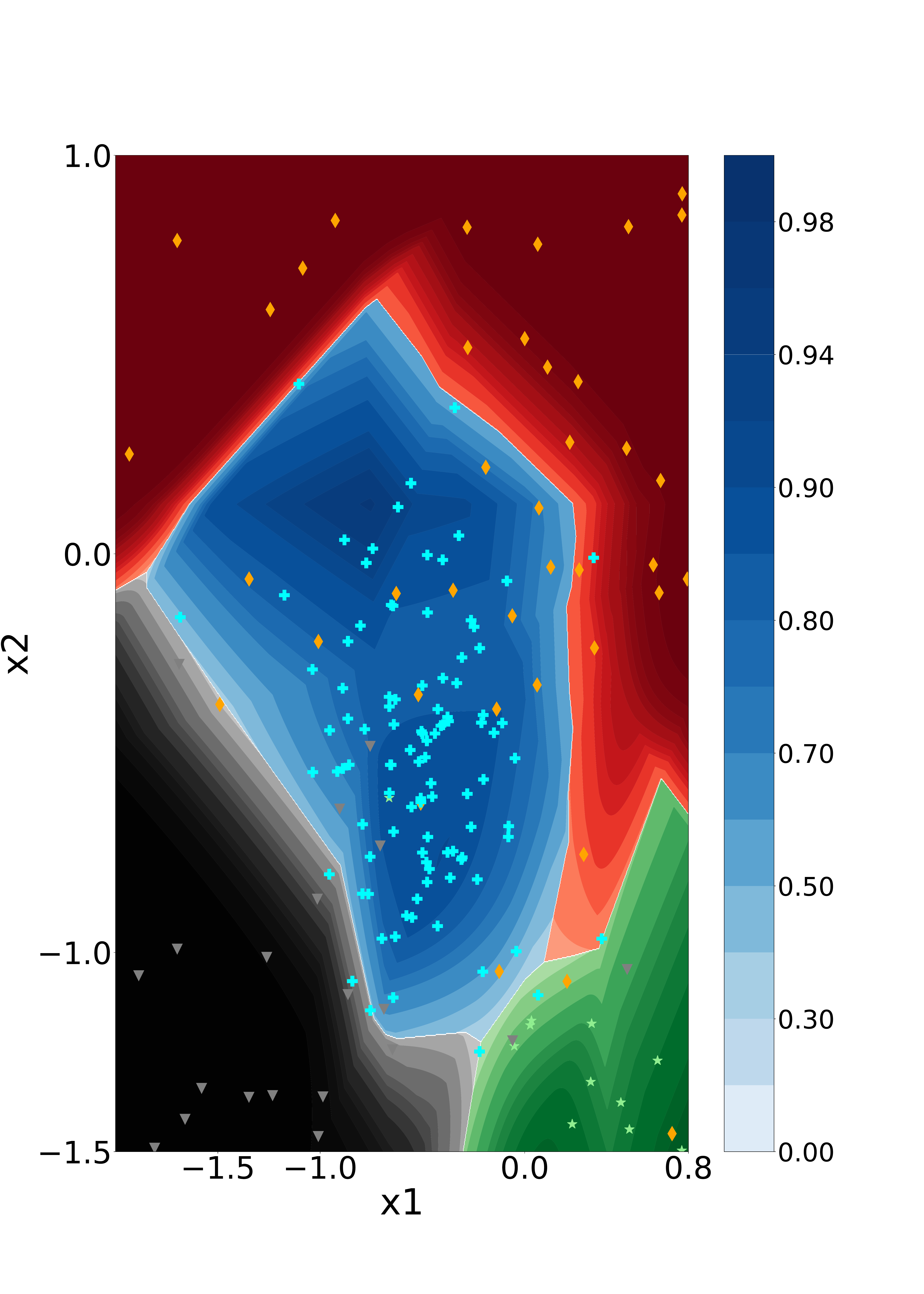}
\end{subfigure}
\caption{Decision thresholds learned by a neural network on a 2-D toy dataset problem where four classes are considered, each one represented with a different colour and marker style. The plot represents the confidence assigned by the model towards the most probable class, in each region of the input space. Darker colours represent higher confidences. The subfigure on the top row left corner represents the decisions learned by a point-estimate model obtained by minimizing the loss function given by equation \ref{equation_MAP}; and the figure on the top row, right corner, represents the confidences learned by a Bayesian model that uses Hamiltonian Monte Carlo to draw samples of the posterior distribution, which are used to approximate the posterior predictive, see \cite{1206.1901} for details. Bottom rows represent zooms to different regions of the input space, showing the decision thresholds learned by the Bayesian model. Each figure represents the Accuracy (ACC) (the higher the better); and the Expected Calibration Error (ECE) (the lower the better). With markers, we plot the observed data $\mathcal{O}$. Figure best viewed in colour. 
}
\label{fig:introduction_figure1}
\end{figure}

To illustrate these claims, figure \ref{fig:introduction_figure1} shows the confidences respectively assigned by Bayesian and point-estimate models based on a neural network (NN) architecture 
in the different parts of the input space, alongside the training data points. The problem consists of a 2-D toy dataset where four classes are considered, each one represented with a different colour. We can see two important aspects. The first one is that the Bayesian model assigns better probabilities, thus being closer to the optimal decision rule. This is reflected by the values of the accuracy and the expected calibration error (ECE) (details on these metrics are provided in the experimental section). Second, it can be seen how the different models assign different confidences on each region of the input space. For the sake of illustration, in the bottom row, we present two different concrete parts of the input space. We can clearly see how the Bayesian model assigns confidence being coherent with what the input distribution presents: highest confidence (close to $1.0$) in regions where only one class is presented and moderate probabilities in regions where the data from different classes overlap. The point-estimate does not present this behaviour. 

Finally, considering likelihood models parameterized by Neural Networks with ReLU activations, one can expect that the predictions made by the Bayesian and Point Estimate approaches do not necessarily converge to the same model as the number of observations tend to infinity, contrary to other simple approaches, e.g. Bayesian linear regression (see \cite{Bishop:2006:PRM:1162264} chapter 3). This means that, even with larger datasets, the predictions done by a BNN can be substantially different from the ones performed by a point estimate one, which justifies the use of Bayesian models in the context of large-scale machine learning. We provide evidence on this observation in the experimental section.

\section{Bayesian Models and Deep Learning}

Having motivated the good properties of the Bayesian reliable probabilistic modelling, in this section we introduce our approach, showing how we overcome many of the limitations that make Bayesian models unpractical when applied to DNNs, and thus how we combine the best of Bayesian inference and deep learning. The approximations presented in this section are motivated by our interest in providing a solution that is both efficient and scalable with dataset size. Therefore, it is expected that much better results will be obtained by using BNNs with more sophisticated approximations, with independence of the pre-trained DNN to calibrate. However, this is outwith the scope of the present work, as our main motivation is providing evidence that the presented approach, a Bayesian stage for recalibration, can consistently improve the calibration. Future work will be concerned with the analysis of different Bayesian stages for this purpose.

\subsection{Proposed Framework}

Our proposal is divided into two steps. First, we train a DNN on a specific task. After training is finished we project each input sample to the logit space, i.e., the pre-softmax, by forwarding the inputs through the DNN. Second, a Bayesian stage is applied, which is responsible for mapping the uncalibrated logit vector of values provided by the DNN, to a calibrated one. Note that once the DNN is trained and the forward step is done for a given sample, the Bayesian stage does not require further access to the previous DNN to be trained, which is why our method is \emph{decoupled}. A graphical depiction is given in figure  \ref{fig:proposed_architechture}.

One should expect this approach to work because of the following reason. DNNs provide high discriminative performance on many complex tasks. However, they overfit the likelihood \cite{DBLP:journals/corr/GuoPSW17}. To correct this uncalibrated probabilistic information, we incorporate a Bayesian stage, which will adjust these confidences, but instead of starting from raw data, it starts from the representation already learned by the DNN in the form of the logit values. As this is a much simpler task than mapping directly the real inputs to class probabilities, we can benefit from the properties of Bayesian inference even though the current state-of-the-art presents many limitations that would not allow us to achieve the same representations learned by a point estimate DNN using the Bayesian counterpart\footnote{Monte Carlo (MC) Dropout \cite{mcdropoutgal} is an exception that will be discussed in the experimental section}. 

We now describe our design choices for the Bayesian stage, which includes the selection of the likelihood and the prior distribution; and the set of approximations derived from these choices.

\subsection{Likelihood Model}
\label{seccion_elegir_likelihood}

In this work, we focus on finite parametric likelihood models $p(t|x,\theta)$, i.e. Bayesian Neural Networks (BNNs), implemented with fully-connected neural networks with ReLU activations for the hidden layers, and a softmax activation for the output layer. Note that one can adapt the complexity and flexibility of this stage depending on the context, for instance by using recurrent architectures.
 
Although Gaussian Processes (GPs) have been recently used for calibration, we discard their study for two reasons. First, their calibration properties depend on the choice of the covariance function \cite{Gal2016Uncertainty}. Second both GPs and BNNs  present similar limitations in a classification context: approximation of the predictive distribution and sampling from (and sometimes approximating) the posterior distribution. However, GPs require additional approximations when dealing with large datasets, e.g. by choosing inducing points \cite{NIPS2005_2857} to parameterize the covariance functions; alongside with heavy matrix computations and huge amounts of memory resources to store data. Moreover, in BNNs inference can be done by simple ancestral sampling, even if we make our models deeper or recurrent; but the current state-of-the-art inference technique in Deep-GPs  \cite{NIPS2018_7979} is based on the Stochastic Gradient Hamiltonian Monte Carlo algorithm \cite{Chen:2014:SGH:3044805.3045080}, which is impractical for the purpose of this work.  
 
\subsection{Inference}
\label{seccion_inference}
In order to predict a label $t^*$ over a new unseen sample $x^*$  we need to compute the expectation described in equation \ref{bayesian_integral}.  The form of the likelihood $p(t|x,\theta)$ as described above makes unfeasible the computation of an analytic solution for the predictive $\hat{p}(t|x)$. Thus, this integral is approximated using a Monte Carlo estimator, given by:

\begin{equation}
    \hat{p}(t^*|x^*) \approx \frac{1}{K} \overset{K}{\underset{k=1}{\sum}}\, p(t^*|x^*,\theta_k);\, \theta_k \sim p(\theta|\mathcal{O})
    \label{montecarlo_predictive}
\end{equation}

As we choose a categorical likelihood $p(t|x,\theta)$, this approximation relies on averaging the softmax output from the different forward steps. In a deep learning context, this likelihood would be a DNN, e.g. a DenseNet-169 \cite{DBLP:journals/corr/HuangLW16a}; and this would require  to perform  $K$ forward steps through it in order to make predictions, which is very costly in terms of computation. However, in our proposed framework, predictions only require one forward step through the DNN, and $K$ forward steps through a much lighter likelihood model. It is worth to say that these predictions are independent and can be totally paralellized. Thus, computational efficiency is not compromised.

\subsection{Sampling from the posterior}

In order to perform inference as described in equation \ref{montecarlo_predictive} we need to draw samples $\theta_k$  from the posterior distribution $p(\theta|\mathcal{O})$, which can be done in two ways. First: by computing an analytic expression or an approximation to the posterior, that will allow us, hopefully, straightforward sampling. Second: using Markov Chain Monte Carlo (MCMC) algorithms that provide exact samples from the posterior without requiring access to it. In this work, we attempt for the first option, as the common MCMC algorithm in BNN, Hamiltonian Monte Carlo (HMC) \cite{1206.1901}, requires careful hyperparameter tuning, among other drawbacks (see \cite{betancourt2017conceptual}). This tuning process has become unfeasible for such an extensive battery of experiments like the one in this work; and thus, it will be only used as an illustrative tool in a toy experiment in the experimental section. 

Based on the choice of the likelihood, the posterior distribution from equation \ref{bayesian_posterior} cannot be computed analytically. For that reason, we approximate this posterior distribution in terms of simple and tractable distribution $q_\phi(\theta) \in \mathcal{Q}$ where $\phi$ denotes the parameters. In order to perform this approximation, we follow a classical procedure in variational inference, by optimizing a bound on the marginal likelihood commonly referred as the Evidence Lower Bound (ELBO) \cite{Bishop:2006:PRM:1162264}, which ensures that the variational distribution is approximated to the intractable posterior $p(\theta|\mathcal{O})$ in terms of the Kullback-Liebler divergence $D_{KL}[q_\phi(\theta)||p(\theta|\mathcal{O})]$. Our choice for the variational distribution family $\mathcal{Q}$ is the factorized Gaussian distribution. The choice of the prior $p(\theta)$ is the standard Gaussian. With this, our training criteria is given by:

\begin{equation}
    q^*_\phi(\theta)=\underset{q_\phi(\theta) \in \mathcal{Q}}{\mbox{argmax}}\,\, M^{-1} \overset{M}{\underset{m=1}{\sum}} \log p(t|x,\theta_m) -\beta D_{KL}[q_\phi(\theta)||p(\theta)];\,\, \theta_m \sim q_\phi(\theta)
    \label{equation:ELBO}
\end{equation}

 where $\beta$ is a hyperparameter controlling the importance provided to the $D_{KL}$. We use the recently proposed reparameterization trick \cite{1312.6114,1401.4082} and the local reparameterization trick \cite{NIPS2015_5666} to allow for unbiased low-variance gradient estimators. We call the first approach as Mean Field Variational Inference (MFVI), and MFVILR (after local reparameterization) to the latter. The motivation below experimenting with these two approaches is made explicitly in the next section. It should be noted that both approximations leave the variational distribution unchanged, i.e. it is still factorized Gaussian. Remark that this approach might be inaccurate and costly to train if applied directly to recover a Bayesian DNN, even if we choose to approximate the posterior distribution using more complex families. However, as supported by our experimental results, it is enough to provide state-of-the-art calibration performance when used under the proposed framework, thus manifesting the ability to combine the best of DNNs and Bayesian modelling.

As a consequence of the choices presented in this section, predictions will be now done by substituting the intractable posterior with the variational approximation. Thus, and after training is finished, the whole pipeline to make a prediction is given by: 
 
 \begin{equation}
 \begin{split}
     \mbox{logit}^* &= \mbox{DNN}(x^*)\\
     \hat{p}(t|\mbox{logit}^*) &\approx \frac{1}{K} \overset{K}{\underset{k=1}{\sum}}\, p(t|\mbox{logit}^*,\theta_k);\, \theta_k \sim q(\theta)\\
     t^* &= \underset{t}{\mbox{argmax}}\,\,\hat{p}(t|\mbox {logit}^*)
 \end{split}    
 \end{equation}

\subsection{Variance Under-Estimation}
\label{variance_overestimation}
One of the drawbacks that this particular Bayesian approximation presents is variance under-estimation (VUE), which is due to the expression of the $D_{KL}$ being minimized as a consequence of optimizing the ELBO (see\cite{Bishop:2006:PRM:1162264} page 469). This makes the variational distribution $q^*_\phi(\theta)$ avoid placing high density over regions where $p(\theta|\mathcal{O})$ presents low density. Or, in other words, if $p(\theta|\mathcal{O})$ is highly multimodal the variational distribution will tend to cover only one mode from the intractable distribution. This effect is also known as mode collapse.

In practice, we realize that this effect affects the performance of the proposed approach in two ways. On one side, consider a highly multimodal intractable posterior that presents a single high-density mode, alongside with different bumps over the parameter space. As a result of the optimization process, if the variational distribution accounts for this high mode, the set of weights sampled could resemble those of MAP estimation, and thus we will be providing over-confidence predictions. To overcome this last limitation, we propose to select the optimal value of $K$ in equation \ref{montecarlo_predictive} on a validation set. While this approach contrasts with the theory, which states that $K$ should tend to infinity, we find it an effective solution to overcome this limitation in our experiments for this particular mean-field approach. 

On the other hand, if our intractable posterior presents several bumps with equal probable density, or our approximate distribution accounts for a non-highly probable mode of the intractable posterior, the set of weights sampled could not be enough representative of the data distribution. The confidences assigned by model parameterized with these set of sampled weights could affect the accuracy and the calibration error. This can only be solved by using more sophisticated approximations of the variational distribution as the MFVI approach can only recover unimodal Gaussian distributions. We realized that this effect only affects the most complex tasks. For complexity, we refer, on one side, to the particular task to solve (which will mainly depend on the number of classes and number of samples) and, on the other to how well the variational distribution is able to fit the intractable posterior. This will depend on the choice of likelihood $p(t|x,\theta)$ and the prior $p(\theta)$; and the set of observations $\mathcal{O}$. Thus, both the number of classes, the representations learned by a DNN and the number of training points play a major role in the final performance of the proposed approach. We will illustrate these claims in the next section. 

\section{Experiments}

We conduct several experiments to illustrate the different properties of the proposed approach. We provide code for reproducibility and supplementary material for details on different results.\footnote{Github: \href{https://github.com/jmaronas/DecoupledBayesianCalibration.pytorch}{https://github.com/jmaronas/DecoupledBayesianCalibration.pytorch}.}.

\subsection{Set-up}

\paragraph{Datasets} We choose datasets with a different number of classes and sizes to analyze the influence of the complexity of the calibration space and the robustness of the model. In parenthesis, we provide the number of classes: Caltech-BIRDS (200)\cite{WahCUB_200_2011}, Standford-CARS (196)\cite{KrauseStarkDengFei-Fei_3DRR2013}, CIFAR100 (100)\cite{cifar100}, CIFAR10 (10)\cite{cifar10}, SVHN (10)\cite{noauthororeditor}, VGGFACE2 (2)\cite{Cao18}, and ADIENCE (2)\cite{Eidinger:2014:AGE:2771306.2772049}. We use all the training set to train the Bayesian models except for VGGFACE, where we use a random subset of 200000 samples, which is 15 times fewer than the original. This was enough to outperform the state-of-the-art.

\paragraph{Models} We evaluate our model on several state-of-the-art configurations of computer vision neural networks, over the mentioned datasets: VGG \cite{vgg_1409.1556}, Residual Networks \cite{DBLP:journals/corr/HeZRS15}, Wide Residual Networks \cite{DBLP:journals/corr/ZagoruykoK16}, Pre-Activation Residual Networks \cite{1603.05027}, Densely Connected Neural Networks \cite{DBLP:journals/corr/HuangLW16a}, Dual Path Networks \cite{dpn_1707.01629}, ResNext \cite{resnext_1611.05431} , MobileNet\cite{Sandler_2018_CVPR} and SeNet \cite{Hu18}.

\paragraph{Performance Measures}In order to evaluate our model, we use the Expected Calibration Error (ECE) \cite{DBLP:journals/corr/GuoPSW17} and the classification accuracy. The ECE is a calibration measure computed as:

\begin{equation}
    \mbox{ECE}=\overset{15}{\underset{i=1}{\sum}} \frac{|B_i|}{N}|\mbox{acc}(B_i) - \mbox{conf}(B_i)|
\end{equation}

\noindent where the $[0,1]$ confidence range is equally divided in bins $B_i$, over which the accuracy $\mbox{acc}(B_i)$ and the average confidence $\mbox{conf}(B_i)$ are computed.

\paragraph{Training specifications} We optimize the ELBO using Adam optimization  \cite{adam_1412.6980} as it performed better than Stochastic Gradient Descent (SGD) in a pilot study, and we select $\beta$ in Equation \ref{equation:ELBO} from the set $\{10^{−i}\}^4_{i=0}$ , depending on the BNN architecture. We use a batch size of 100 and both step and linear learning rate annealing. More details provided in the supplementary material.

\paragraph{Calibration Techniques} We evaluate our model against recently proposed calibration techniques. Regarding explicit techniques, we compare against Temperature Scaling (TS) \cite{DBLP:journals/corr/GuoPSW17} as to our knowledge is the state-of-the-art in decoupled calibration techniques. TS maximizes the log-likelihood of the conditional distribution $p(t|l/T)$ w.r.t the parameter T. $l$ stands for the logit, i.e. pre-softmax of the DNN model (same input as our approach). As all the logits are scaled by the same value, TS is a technique that does not change the accuracy. We also compare with a modified version of Network Ensembles (NE) \cite{NIPS2017_7219}. This is an implicit calibration technique that proposes to average the output of several DNNs with adversarial noise \cite{43405} regularization, different random initialization and randomized training batches. Due to the high computation cost, we train decoupled NE, i.e, NE that maps the logit from the DNN.

On the other hand, regarding implicit calibration techniques, we compare against NE in their original form; and also against MMCE \cite{pmlr-v80-kumar18a}, which proposes a calibration cost which is computed using kernels; and with Monte Carlo Dropout \cite{mcdropoutgal}, that averages several stochastic forward passes through a Neural Network. 

\subsection{Bayesian vs Point Estimate and Variance Under Estimation}

We begin by conducting a series of experiments comparing Bayesian and non-Bayesian approaches using the same toy dataset used in section \ref{BayesianModellingandCalibration}. We aim at illustrating the good calibration properties of the chosen Bayesian model, and its better performance when compared to point-estimate approaches in the presence of bigger training sets. We further illustrate the influence of VUE in the approximate Bayesian model.

We start by evaluating the calibration performance of Bayesian and non-Bayesian models when the number of training samples is large. For this experiment, we use 4000 training samples, which we consider to be a large dataset due to the simplicity of this toy distribution. This toy problem allows using HMC to draw samples from the intractable posterior used to approximate the predictive distribution in the Bayesian model. For the point estimate, we use a MAP training criteria optimized with SGD and momentum. Results are shown in table \ref{tab:HMC_MFVILR_MAP}, where we compare different induced posterior distributions showing how the calibration error of the Bayesian HMC model is one order of magnitude below the point estimate MAP. Thus, one should expect that for more complex distributions than this of our toy dataset will be further improved by a Bayesian approach.

\begin{table}[!t]
    \centering
    \caption{A comparison between HMC MFVILR and MAP using 4000 training samples. Prior specifies prior variance. Likelihood specifies hidden-layers/neurons-per-layer}
    \label{tab:HMC_MFVILR_MAP}
    \scalebox{0.7}{
    \begin{tabular}{c c | c c  c c c c }
    \multicolumn{2}{c|}{posterior specs} & \multicolumn{2}{c}{HMC} & \multicolumn{2}{c}{MFVILR}& \multicolumn{2}{c}{MAP}\\\hline
    prior & Likelihood & ACC & ECE & ACC & ECE & ACC & ECE\\\hline
      16   &     0/-      & 85  & 0.05  & 61.0 &  0.25 & 83 & 0.29 \\\hline
      16   &     1/25     & 86 & 0.05 & 67.0 & 0.19 & 85 & 0.26 \\\hline
      16   &     1/50     & 86.5  & 0.05 & 67 & 0.21 & 84.5 & 0.26 \\\hline
      32   &     0/-      & 85  & 0.05  & 66.0 &  0.23 & 86 & 0.26 \\\hline
      32   &     1/25     & 87  & 0.04  & 79.5 & 0.19 & 85.5 & 0.19 \\\hline
      32   &     1/50     & 86.5  & 0.05  & 81.0 & 0.22 & 86 & 0.18 \\\hline
\end{tabular}}
\end{table}

We then illustrate the effect of variance under-estimation (VUE). As we argued above, in the context of BNNs for classification, this VUE effect can cause accuracy degradation and bad calibrated predictions. Using the results from table \ref{tab:HMC_MFVILR_MAP} we compare the performance of the Bayesian model using HMC and MFVILR. As expected, MFVILR is providing worse calibration and accuracy than HMC, clearly due to a bad approximation to the intractable posterior. We can further highlight this effect by taking a look at the 0-hidden layer likelihood model. Under this parameterization, the intractable posterior is a non-Gaussian unimodal distribution and, even though our approximation is also unimodal, it cannot correctly fit the intractable posterior.

\subsection{Bayesian vs Non-Bayesian Linear Regression}
\label{bayesianvsnonbayesianlinearrergession}
In this section, we compare Bayesian and non-Bayesian Linear Logistic Regression under the proposed framework. We train several DNNs on different datasets and then use a Linear Logistic model with a Bayesian and a Non-Bayesian approximation. In this setting, the likelihood is given by:

\begin{equation}
    p(t|x,\theta) = f(x^T\cdot W+b),
\end{equation}
where $W$ and $b$ are parameters, $f()$ is the softmax function and $x$ represents the logit computed from the DNN. 

The motivation below this comparison is based on the observation that, as shown in table \ref{tab:HMC_MFVILR_MAP}, one could think that our approach (MFVILR) provide worse results than a point estimate model. However, as we now show, when combined with a DNN it outperforms the point estimate approach. Moreover, we want to show that the poor calibration capabilities of complex techniques, as strengthened by \cite{DBLP:journals/corr/GuoPSW17}, are due to bad treatment of uncertainty, and not because the calibration space is inherently simple. 

Table \ref{tabla:bayesian_linear_regression} shows a comparison of both methods where it is clear that the Bayesian model provides better performance both in accuracy and calibration. It should be noted that the solution of this optimization problem under the non-Bayesian estimation is unique, while the MFVILR admits several steps of improvement just by using more sophisticated approximated distribution, that could capture non-Gaussian or multimodal posteriors. Thus, it is clear that our main claim, combining the powerfulness of DNNs and BNNs can be achieved. 

\begin{table}[!t]
 \caption{Calibration ECE (\%), and accuracy (ACC) (\%) performance  for averages of several logistic models trained for three of the databases considered in this work. ACC the higher the better, ECE the lower the better. \newline}
    \centering
    \scalebox{0.8}{
    \begin{tabular}{ c c c c c c c}
    & \multicolumn{2}{c}{\textbf{CIFAR100}} & \multicolumn{2}{c}{\textbf{SVHN}} & \multicolumn{2}{c}{\textbf{CARS}}\\
        & ECE & ACC & ECE & ACC & ECE & ACC \\ \hline
    Point Estimate & 33.90 & 62.67 & 1.13 & 96.72 & 23.50 & \textbf{76.14}\\
    Bayesian    &  \textbf{3.66} & \textbf{72.36} & \textbf{1.03} & \textbf{96.72} & \textbf{1.88} &  74.31 
    \end{tabular}}
    \label{tabla:bayesian_linear_regression}
\end{table}

\subsection{Selecting optimal on validation}

We then illustrate why selecting the optimal value of Monte Carlo predictive samples with a validation set is necessary. One of the problems of VUE is that we can fit our approximation to a high-probable mode of the intractable posterior density, sampling set of weights that could resemble those of MAP estimation, with overconfidence probability estimates as a result. In this work we show that this effect can be controlled by searching for the optimal value of Monte Carlo predictive samples,  $K$ in equation \ref{montecarlo_predictive}, using a validation set. 

\begin{figure}[!b]
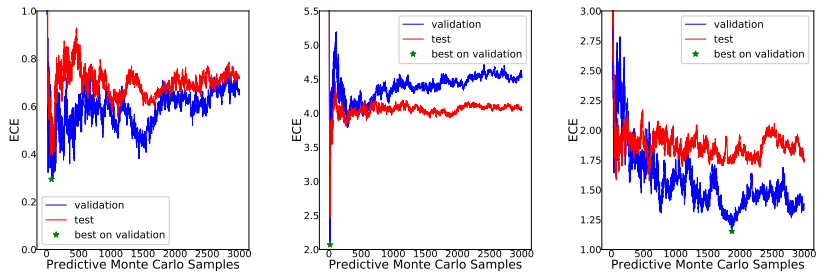

    \centering
    \begin{subfigure}[l]{0.3\columnwidth}
      \includegraphics[width=1.0\columnwidth]{figures/figure5/voverest_cifar10_wrn40x10.pdf}
    \end{subfigure}
    \begin{subfigure}[c]{0.3\columnwidth}
\includegraphics[width=1.0\columnwidth]{figures/figure5/voverest_cifar100_dn121.pdf}
    \end{subfigure}
    \begin{subfigure}[r]{0.3\columnwidth}
      \includegraphics[width=1.0\columnwidth]{figures/figure5/voverest_cifar100_rn101.pdf}
    \end{subfigure}
    \caption{ECE measure on validation and test set varying the number of Monte Carlo Predictive samples. From left to right: cifar10 WideResnet-40x10, cifar100 DenseNet-121, cifar100 ResNet-101.}
    \label{fig:selectKvalidation}
\end{figure}

As an illustration of this over-sampling effect, figure \ref{fig:selectKvalidation} shows the calibration error when increasing the number of MC samples. By looking at the figure in the middle and in the left we can see how the calibration error is kept constant (or even increased) when more samples are drawn. This suggests that the variational distribution is coupling to a particular part of the intractable posterior. As a consequence, the ultimate confidence assigned by the model is not being consistent with the ideal estimation. In the case of being coupled to high probability regions of the intractable posterior, the generated samples could resemble those of map estimation, having overconfidence predictions as a consequence, which links with the observations provided by \cite{DBLP:journals/corr/GuoPSW17} in which complex models provide overconfidence predictions. However, this effect can be more or less present, as seen for instance in the right figure, where the behaviour resembles what one should expect, i.e. better performance when increasing the number of MC samples.  However, even without selecting for the optimal value of $K$ on validation, we observed that most of the models outperformed the baseline uncalibrated DNN and provide competitive or even better results than the state-of-the-art as $K$ increases.

\subsection{Calibration performance of BNNs}
 In this subsection, we discuss the calibration performance of the proposed framework. We start by evaluating the proposed method against a baseline uncalibrated network several datasets. Results are shown in table  \ref{decoupled_calibration_results}, where we compare the results with MFVILR and MFVI. For VGGFACE2 we only run the experiments with MFVILR due to computational restrictions.

\begin{table}[!t]
\centering
\caption{Average ECE 15(\%) and ACC(\%) on the test set comparing the uncalibrated model, and the model calibrated with MFVI and MFVILR for each database. ECE15 the lower the better, ACC the higher the better. "degr" means degraded\newline}
\label{table:calibration_results}
\scalebox{0.8}{
\begin{tabular}{ccccccc}
                         & \multicolumn{2}{c}{uncalibrated}&  \multicolumn{2}{c}{MFVI} & \multicolumn{2}{c}{MFVILR}\\ \cmidrule(r){2-7}
                         & Acc             & ECE  & Acc             & ECE            & Acc                & ECE            \\ \cmidrule(r){1-7}
CIFAR10             &     94.81       &    3.19      &  94.70  &  0.58   & 94.64  &    \textbf{0.50}  \\
SVHN          &     96.59       &    1.35      &  96.50   &  0.87    &   96.55 &     \textbf{0.85}  \\
CIFAR100              &     76.36       &    11.39      &  73.87  &     2.52 &   74.44 &     \textbf{2.52}  \\
VGGFACE2 &     96.19       &    1.33      &    -    &  -  &  96.20 & \textbf{0.37} \\
ADIENCE              &  94.25  &  4.55 &  94.28 &  0.53 & 94.27 &  \textbf{0.51}      \\
BIRDS             &     76.27      &   13.22   & degr  & degr  &  74.32 &  \textbf{1.88}  \\
CARS             &    88.79       &    5.81    & degr  & degr  &  85.34 &  \textbf{ 1.59}    \\
\hline\hline  
\end{tabular}}
\label{decoupled_calibration_results}
\end{table}
As shown in the table, the proposed technique improves the calibration performance by a wide margin over the baseline even though we are using a mean-field approximation to the intractable posterior distribution with well-known established limitations. Regarding the accuracy performance, we see a slight accuracy degradation which is only relevant in highly complex tasks, such as CIFAR100, BIRDS and CARS. Our hypothesis is that this degradation is not due to a limitation of the BNN algorithm, but due to inaccurate approximations to the true posterior in some settings. In fact, in some cases, we improve the accuracy over the baseline, as in the two-class problem. This degradation can also give us further insight into the complexity of the calibration task. 

As we stated, accuracy degradation can be explained by mode collapse. To illustrate this claim, we compare the performance provided by MFVI and MFVILR, as both these approximations only differ in the convergence rate of the training criteria from equation \ref{equation:ELBO}, i.e, both approximations provide factorized Gaussian approximations $q_\phi(\theta)$ as approximate distributions.  As shown by the table, better results were obtained by the MFVILR, both regarding calibration and accuracy performance, which means that an inaccurate approximation to the true posterior is responsible for this degradation.  This is justified by the fact that, as the MFVILR provides better convergence rate, we are able to fit a better approximation to the intractable posterior. This same effect is showed when one trains the same DNN using SGD and SGD with momentum. Even the models and the initialization can be the same, the results provided by SGD with momentum are better due to the lower noisy gradients. 

On the other hand, as we see from the results, this degradation is noticeable in more complex tasks. This suggests that the complexity of the intractable posterior increases with the complexity of the task, and thus, a mean-field approximation is not able to provide the same performance as it does in simpler ones. It should be noted that more complex decision regions will induce more complex posteriors, through the likelihood term in equation \ref{bayesian_posterior}. This follows our claim that complex techniques overfit due to a bad uncertainty treatment and not because the calibration space is inherently simple, as noted in \cite{DBLP:journals/corr/GuoPSW17}. To provide further insight,  table \ref{localrep_vs_nonlocalrep} compares MFVI and MFVILR with different models and CIFAR100. The first two rows of the table show how the accuracy degradation is clearly improved just by using MFVILR, which is a general tendency in the experiments (see the supplementary material). However, one can not expect that using MFVILR should always achieve better results, as a good convergence of MFVI should make us recover similar approximate posteriors, reflected as no performance increases.  This is shown in the third and fourth rows. Moreover, if the approximate posterior is a bad approximation to the true posterior, we can dig into an undesirable local minimum, as shown in the fifth and sixth rows. We found that models where MFVILR worsened the performance w.r.t MVFI where those more difficult to calibrate in general, which can be explained by the fact that the complexity of the true posterior cannot be captured by the factorized Gaussian approximation, and more sophisticated approximations need to be employed.

\begin{table}[!t]
\caption{MVFI compared to MVFILR in CIFAR100. * means best model on validation}
\label{localrep_vs_nonlocalrep}
\centering
\resizebox{!}{0.12\textheight}{
\begin{tabular}{ccccc} 
                         & \multicolumn{4}{c}{\textbf{CIFAR100}}  \\
                         &  \multicolumn{2}{c}{MVFI} & \multicolumn{2}{c}{MVFILR} \\ \cmidrule(r){2-5}
                         & ACC                   & ECE  & ACC                   & ECE                  \\ \cmidrule(r){1-5}
DenseNet 169             &  75.58 & 2.39  & 77.22* &  2.45  \\
ResNet 101               &  68.59 & 1.61  & 70.31* &  1.75  \\\hline
Wide ResNet 40x10        &  76.17 & 1.88  & 76.51* &  1.79  \\
Preactivation ResNet 18  &  74.30 & 1.76  & 74.51* &  1.59  \\\hline
Preactivation ResNet 164 &  70.77* & 1.46  & 71.16 &  2.20  \\
ResNext 29\_8x16         &  73.97* & 2.58  & 71.13 &  3.77   \\
\hline \hline
\end{tabular}}
\label{tab:overfit}
\end{table}

On the other hand, we can also provide evidence on the complexity of the calibration space as being dependent on the complexity of the task by analyzing another effect observed in the experiments carried out. Again, and only in complex tasks: CIFAR100, BIRDS and CARS, we experimented an accuracy degradation during training with the MFVI. This means that even although the ELBO was correctly maximized, i.e. the likelihood correctly increases over the course of learning, the accuracy provided was totally degraded. In CIFAR100 we solve it by progressively increasing the expressiveness of the likelihood model for the MFVI, as illustrated in the supplementary material. However, on BIRDS and CARS it could only be solved when using MFVILR, as shown in table \ref{decoupled_calibration_results} where "degr" stands for degradation, and it refers to this effect. This suggests that the factorized Gaussian is unable to give a reasonable approximation to the intractable posterior under noisier gradients.  As this effect is only present in a more complex task, this again suggests that when the complexity of the task increases, so does it the calibration space.

\begin{table}[!t]
\centering
\caption{Average number of parameters (in  thousands).}
\label{numberofparams}
\scalebox{0.8}{
\begin{tabular}{ccc}
& MFVI & MFVILR\\\hline
CIFAR100 & 24018.7 & 430.5\\
CIFAR10 & 696.6 & 65.6 \\
SVHN & 606.9 & 7.6 \\
ADIENCE &0.470 & 4.482\\\hline
average & 6331.2 & 126.1\\
\hline\hline  
\end{tabular}}
\end{table}

On the other hand and based on the previous observation, one could argue that accuracy degradation is due to a lack of expressiveness in the likelihood model. However, we still emphasize that VUE is responsible for this effect. This is because first increasing the expressiveness of the likelihood model in MFVI on BIRDS and CARS did not solve the problem. Second is because we observed that by using MFVILR we were able to reduce the topologies, in general, of the likelihood model as compared with MFVI. This is illustrated in table \ref{numberofparams} where we show a comparison between the average number of parameters used for each task \footnote{In ADIENCE MFVILR was not able to reduce the topologies due to instabilities when computing derivatives. We provide a justification in the supplementary material}.

To end with, we surprisingly found that in some models that achieved good calibration and accuracy properties, both the negative-log-likelihood and the accuracy increased over the course of learning. This means that the network is unable to correctly raise the probability toward the correct class for the miss-classified samples. 

\subsection{Comparison Against state-of-the-art calibration techniques}

We then compare the calibration performance of our method against other proposed techniques for calibration, both implicit and explicit. For the comparison, we use the hyperparameters as provided in the original works. Results are shown in table \ref{against_explicit_methods} for explicit methods and in \ref{against_implicit_methods} for implicit methods. Results on the same dataset might differ as due to the high computational cost of some of the explicit calibration techniques, we only perform a subset of the experiments. Details on the models used to compute these results are provided in the supplementary material.

\subsubsection{Explicit calibration techniques}

\begin{table}[!t]
\centering
\caption{Average ECE results compared against explicit calibration techniques.}
\label{against_explicit_methods}
\resizebox{\textwidth}{!}{
\begin{tabular}{cccccccc}
                         & CIFAR10 & CIFAR100 & SVHN & BIRDS & CARS & VGGFACE2 & ADIENCE \\ \cmidrule(r){2-8}
NE decoupled & 2.55 & 10.17 & 1.02 & 5.25 & 5.51 & 0.79 & 2.64 \\
TS \cite{DBLP:journals/corr/GuoPSW17} & 0.90 & 3.29 & 1.04 & 2.41 & 1.80 & 0.55 & 0.87 \\
ours    & \textbf{0.50} & \textbf{2.52} & \textbf{0.85} & \textbf{1.88} & \textbf{1.59} & \textbf{0.37} & \textbf{0.51}\\
\hline\hline  
\end{tabular}}
\end{table}

Comparing against explicit calibration techniques we first see that all the methods increase the calibration performance over the baseline (see table \ref{table:calibration_results}), with a clear improvement of the BNNs over the rest in all the tasks. These results demonstrate the two main hypotheses of this work: Bayesian statistics provide more reliable probabilities, and complex models improve calibration over simple ones. This observation is consistent in all the experiments presented, where the ECE is the lowest for the proposed model, manifesting the robustness of the BNN approach in terms of calibration. Therefore, our results support the hypothesis that point-estimate complex approaches for re-calibration overfit \cite{DBLP:journals/corr/GuoPSW17} because uncertainty is not incorporated and not because calibration is inherently a simple task. This conclusion can also be supported by the fact that as the complexity of the task increases, the number of parameters of the Bayesian model that yields better results also increases. For instance, the calibration BNN for CIFAR100 needs much more parameters than the BNNs for simpler tasks such as CIFAR10, as shown in table \ref{numberofparams}.  Second, it is important to remark that in some models TS has degraded calibration by a factor of three in the worst case while BNNs do not, as seen in the results provided in the supplementary material. On the other hand, Bayesian model average clearly outperforms standard model averaging as performed by NE. In fact, NE are not suitable for the calibration of deep models, because training directly an ensemble of DNNs is computationally hard and training NE over the logit space does not perform as well as TS. In addition, NE is the one that uses more parameters.

 All these observations manifest the suitability of the proposed decoupled Bayesian stage for recalibration, as even a mean-field approximation to the intractable posterior performs better in terms of calibration than the state-of-the-art in many scenarios. This motivates future work to study more complex variational approximations and different Bayesian-based stages, in order to mitigate the accuracy degradation observed in these experiments.

\begin{figure}[!b]
    \centering
    \begin{subfigure}[l]{0.49\columnwidth}
    \caption*{DenseNet-121 CIFAR10}
  \includegraphics[width=0.9\columnwidth,height=4.5cm]{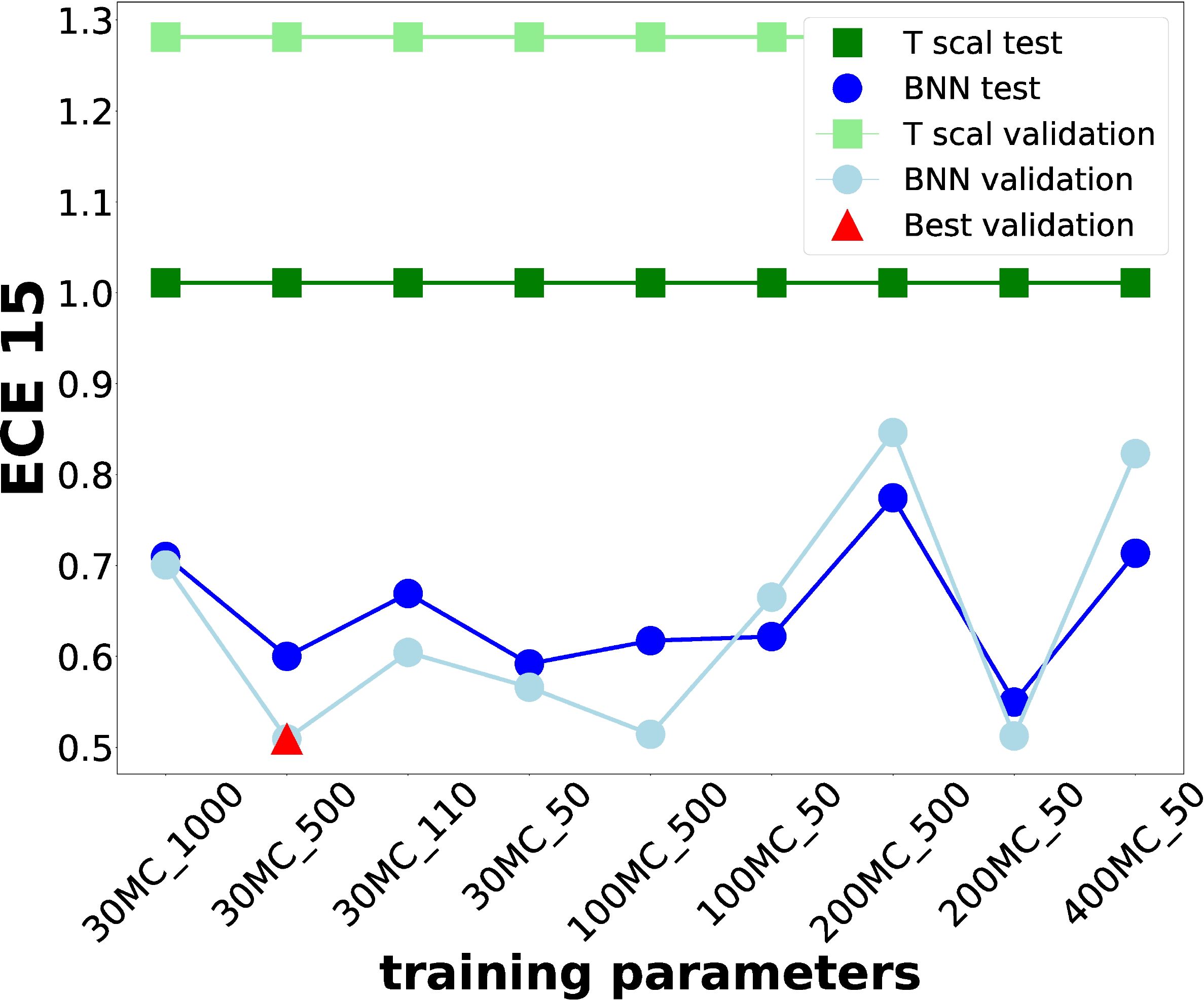}
    \end{subfigure}
    \begin{subfigure}[c]{0.49\columnwidth}
    \caption*{DenseNet-121 CIFAR100}
  \includegraphics[width=0.9\columnwidth,height=4.5cm]{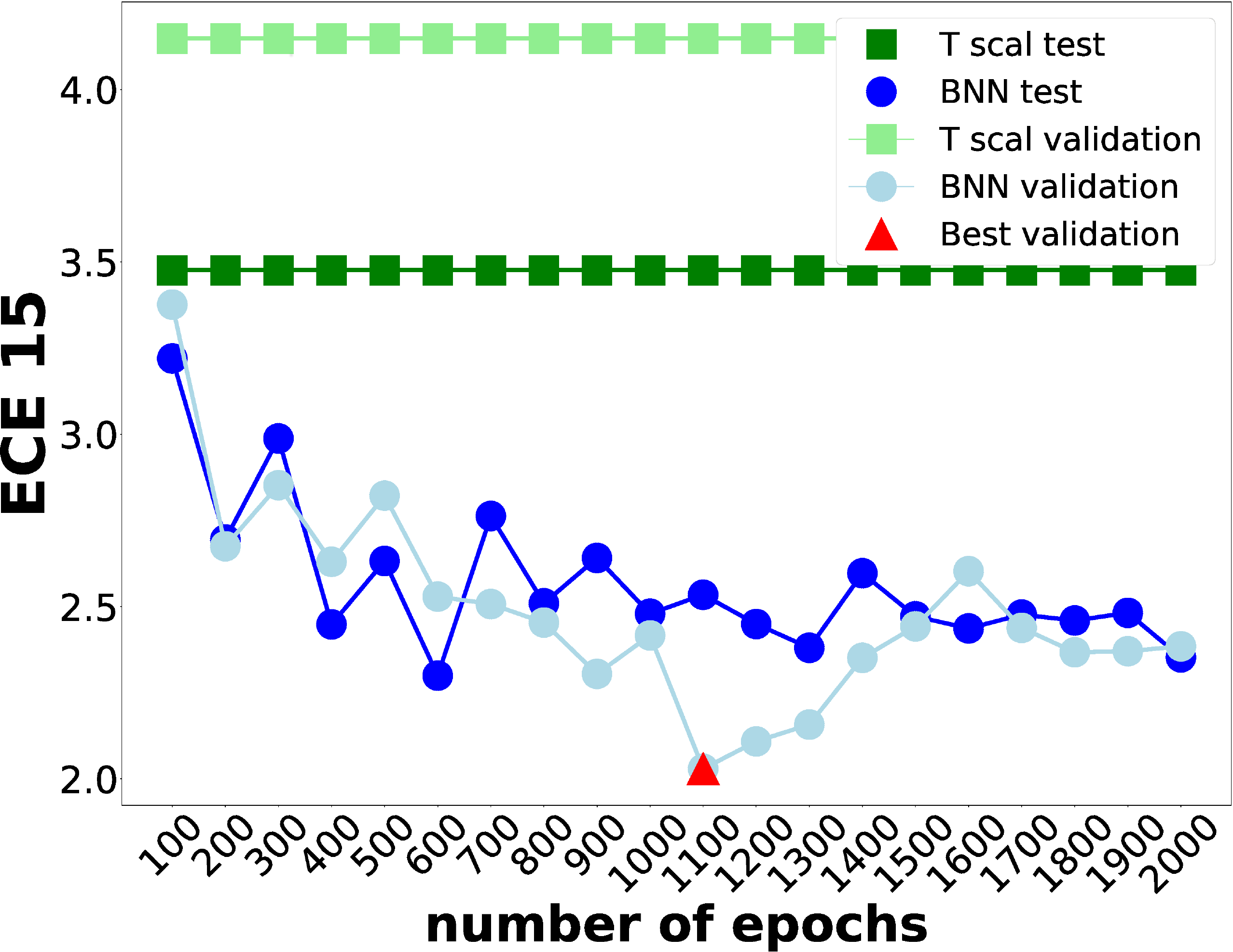}    
    \end{subfigure}
    \caption{Comparison of ECE performance between TS and BNN in test and validation. On the left (CIFAR10) we show the performance of models trained with different parameters. As an example, 30MC 500 means that the ELBO was optimized using 30 MC samples to estimate expectation under $q_\phi(\theta)$ and 500 epochs of Adam optimization. On the right (CIFAR100) we show the performance of a BNN trained with a different number of epochs up to 2000, showing the performance against the course of learning. The number of samples to evaluate the predicted is chosen on a validation set to avoid variance under-estimation.}
    \label{fig:robustness}
\end{figure}

To end with, one important aspect we observed is the robustness of BNNs. We obtained a calibration improvement over TS on the first hyperparameter search in many of the experiments performed. Only some exceptions require further hyperparameter search, which is explained by having to approximate more complex posterior distributions. However, in general, the mean-field approach provides good results, as illustrated in figure  \ref{fig:robustness}, where we show how many of the tested configurations outperformed TS. More figures are provided in the supplementary material. 

\subsubsection{Implicit calibration techniques}

We then compare against implicit calibration techniques. Looking at the results in table \ref{against_implicit_methods} we see that Network Ensembles provide competitive results but at a higher computational cost. This is because this method requires to train several DNN to search for the optimal parameters (number of ensembles, the factor of adversarial noise, topologies of the ensembles...), while we only require to reach good discrimination as provided by the DNN, and then search hyperparameters on a much lighter model.

\begin{table}[!t]
\centering
\caption{Average ECE results compared against implicit calibration techniques. * indicates that the results are taken from the original works. We also include TS. Results from TS and our approach differ from table \ref{against_explicit_methods} as we only pick the DNNs used in the explicit techniques.}
\label{against_implicit_methods}
\resizebox{!}{0.12\textheight}{
\begin{tabular}{cccc}
                         & CIFAR10 & CIFAR100 & SVHN \\ \cmidrule(r){2-4}
VWCI \cite{1809.10877}*  & - & 4.90  & -\\
MMCE \cite{pmlr-v80-kumar18a} & 1.79 & 6.72 &  1.12 \\
TS \cite{DBLP:journals/corr/GuoPSW17} & 0.82 & 3.84 & 1.11 \\
MCDROP \cite{mcdropoutgal} & 1.38 & 3.49 & 0.92  \\
NE \cite{NIPS2017_7219}  & 0.61 & 3.27 &  \textbf{0.71}\\
ours    & \textbf{0.43} & \textbf{2.28} & 0.83 \\
\hline\hline  
\end{tabular}}
\end{table}

On the other hand, we briefly discuss other potential advantages of our method against implicit techniques.  First, we see how our  Bayesian method outperforms the other Bayesian method provided, named Monte Carlo dropout (MCDROP). We should expect these results as the main authors clearly state in their work that the probabilities provided by this method should not be necessarily calibrated as the dropout parameter has to be adapted as a variational parameter depending on the data at hand \cite{NIPS2017_6949}. In fact, many works that aim at reporting that Bayesian methods do not provide calibrated outputs  \cite{NIPS2017_7219,DBLP:conf/icml/KuleshovFE18} only provide results comparing with this technique. However, this work has clearly shown that Bayesian methods are able to improve the calibration performance over point estimate techniques.

Moreover, while our method does not compromise the previous DNN architecture, both MC dropout and VWCI require sampling-based stages, e.g dropout, to be applied to the DNN.  Despite the improvement of \cite{1809.10877} over a baseline uncalibrated model, our method is clearly better, as shown in the table. Moreover, it seems unclear how scalable this method is when applied to Deep Learning models, as to compute the cost function, this approach requires several forwards through the DNN. While their deeper model is a DenseNet-40 we provide results here for a DenseNet-169. On the other hand, our method is clearly more efficient than MC dropout or other Bayesian implicit methods \cite{1805.10522,1805.10915} as these requires performing several forwards through the DNN.

Finally, developing techniques to recalibrate the outputs of a model is indeed interesting, as they can be combined with implicit techniques. As an example, the best results reported by \cite{pmlr-v80-kumar18a} are a combination with their method with TS. Furthermore, \cite{Lee2017TrainingCC} also uses TS as the calibration technique, and \cite{DBLP:conf/icml/KuleshovFE18} proposes a method for re-calibrating outputs in regression problems; which manifest the interest and power of developing techniques that aim at re-calibrating outputs of a model.

\subsection{Qualitative Analysis}

We have also performed a qualitative analysis of the output of the Bayesian model in comparison with TS. We realized that on the misclassified samples made by TS and BNNs, the BNN assigns lower confidence than TS, which is a desirable property. On the other hand, regarding the correctly classified samples, the BNN not only adjusts the confidence better but also classifies these samples with higher confidence than TS. This may mean than TS calibrates by pushing samples to lower confidence regions, an observation that has been also noted in previous works \cite{pmlr-v80-kumar18a}. Moreover, we analyzed the samples where the BNN decided a different class w.r.t the DNN. On the one hand, we analyzed the set of these samples where the class assigned by the BNN was correct, i.e. 100\% accuracy. First, in this set, the original decision made by the DNN was incorrect, i.e. 0\% accuracy. Second, the DNN assigned very high incorrect confidence (over 0.9) to some of these miss-classified samples. Third, the new confidence assigned by the BNN was not extreme, which means that the BNN “carefully” changes the decision made by the DNN. On the other hand, we analyze the set of samples where the BNN assigned a different class from the DNN, and this newly assigned class was incorrect. First, we realize that the DNN only had a 50\% of accuracy on this set. Second, the original confidence assigned by the DNN to these samples was below 0.5. This means that the BNN does not make wrong decisions on a set of high-confidence, well-classified samples by the DNN. 

\section{Discussion}

Having presented and evaluated the proposed approach, here we enumerate and summarize a number of their advantages and lines of improvement. First, the Bayesian stage is only compromised by the dimensionality of the logit space, no matter how challenging the initial task is, or the type and complexity of the pre-trained DNN. Second, the approach is efficient, since the initial DNN model does not need to be re-trained for re-calibration. Some approaches that attempt to directly train a deep calibrated model \cite{pmlr-v80-kumar18a,1809.10877} increase the training time over the initial DNN. In this sense, hyperparameter search is quicker with our proposal, as we only need to focus on getting good accuracy from the DNN. Third, we can incorporate future improvements to the BNN calibration stage without affecting the previous DNN model. For instance, recent proposals such as \cite{fixing} or Bayesian stages based on Gaussian processes \cite{NIPS2018_7979}. Fourth, our proposal is extremely flexible, as the proposed BNN calibration stage will work with any probabilistic model, including models that are designed to be implicitly calibrated  \cite{pmlr-v80-kumar18a,1809.10877}, with potential additional benefits on calibration performance. For instance, the best results reported by \cite{pmlr-v80-kumar18a} are a combination of their method with TS. Fifth, we do not compromise the architecture of the previous stage. Other proposals that attempt to calibrate implicitly \cite{1809.10877}, or to model uncertainty in a Bayesian way \cite{mcdropoutgal}, require certain architectures in the previous stage. Finally, we will show that our approximation is robust, i.e, we provide below better calibration than the current state-of-the-art in many different configurations of the BNNs and optimization hyperparameters.

On the other hand, the disadvantages discussed in section \ref{variance_overestimation} are not a limitation of our approach. We can still improve the approximate posterior by applying normalizing flows \citep{1505.05770,NIPS2016_6581,Huang2018NeuralAF,Berg2018SylvesterNF}, auxiliary variables \citep{DBLP:conf/iconip/AgakovB04a,1511.02386,Maaloe:2016:ADG:3045390.3045543}, combinations of all of them \citep{1703.01961} or deterministic models \citep{fixing}. Also, \cite{pmlr-v80-cremer18a} has recently pointed out that amortized inference leads to an additional gap in the bound, in addition to the $D_{KL}$ gap between the true and variational posteriors; and we can also use other proposals to mitigate this effect \citep{DBLP:journals/corr/abs-1805-08913,pmlr-v80-kim18e}. Finally a potential line of research considers robustification by means of Generalized Variational Inference \cite{knoblauch2019generalized}. However, including all these improvements is not the aim of this work, but to show the adequacy of the proposed decoupled BNN and its potential for future improvements. This is because the true posterior distribution can be highly variable, as it not only depends on the parameterization of the likelihood model and the prior but also on the observed dataset, which itself depends on the input training distribution and the set of representations learned by the specific DNN. Thus we decided to validate our proposal restricting ourselves to the Gaussian approximation and to show it works in a numerous set of different configurations. 

\section{Conclusions and Future Work}

This work has shown that Bayesian Neural Networks with mean-field variational approximations can robustly provide state-of-the-art calibration performance in Deep Learning frameworks, overcoming the limitations of applying Bayesian techniques directly to them. This suggests that using more sophisticated approximations to the intractable posterior should even yield better results than the ones reported in this work. 

We have also shown that as long as uncertainty is properly addressed we can make use of complex models that do not overfit, showing that probability assignments of DNN outputs suppose a more complex task than what previous work argued. Also, we have shown that, in contrast to previous work, Bayesian models parameterized with Neural Networks can be successfully used for the task of calibration. Moreover, our approach is a clear alternative to the development of Bayesian techniques directly applied to DNN, such as concrete dropout\cite{NIPS2017_6949}, as we do it at a much lower computational cost. 

On the other hand, we have analyzed and justified the drawbacks found in this work: slight accuracy degradation in complex tasks and the selection of the number of Monte Carlo predictive samples using a validation set. Future work will be focused on the exploration and analysis of different Bayesian models for the task of calibration, and different approximations to the intractable posterior distribution. With all this, we aim at reducing and deeply analyze the influence of the aforementioned drawbacks. 

\section{Acknowledgement}

We gratefully acknowledge the feedback provided by Emilio Granell and Enrique Vidal on an earlier manuscript. We also acknowledge the support of NVIDIA by providing two GPU Titan XP from their grant program and Mario Parreño for providing the logits of the ADIENCE and VGGFACE2 models. Juan Maro\~nas is supported by grant FPI-UPV.


\begin{thebibliography}{10}
\expandafter\ifx\csname url\endcsname\relax
  \def\url#1{\texttt{#1}}\fi
\expandafter\ifx\csname urlprefix\endcsname\relax\def\urlprefix{URL }\fi
\expandafter\ifx\csname href\endcsname\relax
  \def\href#1#2{#2} \def\path#1{#1}\fi

\bibitem{DBLP:journals/corr/HuangLW16a}
G.~Huang, et~al., Densely connected convolutional networks, 2017 IEEE
  Conference on Computer Vision and Pattern Recognition (CVPR) (2017)
  2261--2269.

\bibitem{DBLP:journals/corr/ZagoruykoK16}
S.~Zagoruyko, et~al., Wide residual networks, in: E.~R.~H. Richard C.~Wilson,
  W.~A.~P. Smith (Eds.), Proceedings of the British Machine Vision Conference
  (BMVC), BMVA Press, 2016, pp. 87.1--87.12.
\newblock \href {http://dx.doi.org/10.5244/C.30.87}
  {\path{doi:10.5244/C.30.87}}.

\bibitem{DBLP:journals/corr/abs-1301-3781}
T.~Mikolov, et~al., Efficient estimation of word representations in vector
  space, in: International Conference on Learning Representations, 2013.

\bibitem{DBLP:journals/corr/MikolovSCCD13}
T.~Mikolov, et~al., Distributed representations of words and phrases and their
  compositionality, in: Proceedings of the 26th International Conference on
  Neural Information Processing Systems - Volume 2, NIPS'13, Curran Associates
  Inc., USA, 2013, pp. 3111--3119.

\bibitem{DBLP:journals/corr/VaswaniSPUJGKP17}
A.~Vaswani, et~al., Attention is all you need, in: I.~Guyon, U.~V. Luxburg,
  S.~Bengio, H.~Wallach, R.~Fergus, S.~Vishwanathan, R.~Garnett (Eds.),
  Advances in Neural Information Processing Systems 30, Curran Associates,
  Inc., 2017, pp. 5998--6008.

\bibitem{hinton16speechprocessing}
G.~Hinton, et~al., Deep neural networks for acoustic modelling in speech
  recognition. the shared views of four research groups, IEEE Signal Processing
  Magazine 29~(6) (2012) 82--97.
\newblock \href {http://dx.doi.org/10.1109/MSP.2012.2205597}
  {\path{doi:10.1109/MSP.2012.2205597}}.

\bibitem{dawid82wellCalibratedBayesian}
A.~P. Dawid, The well-calibrated {Bayesian}, Journal of the American
  Statistical Association 77~(379) (1982) 605--610.

\bibitem{cohen04calibrated}
I.~Cohen, et~al., Properties and benefits of calibrated classifiers, in:
  Knowledge Discovery in Databases: PKDD 2004, Vol. 3202 of {Lecture Notes in
  Computer Science}, Springer, Heidelberg - Berlin, 2004.

\bibitem{brummer10PhD}
N.~Br\"ummer, Measuring, refining and calibrating speaker and language
  information extracted from speech, Ph.D. thesis, School of Electrical
  Engineering, University of Stellenbosch, Stellenbosch, South Africa,
  available at http://sites.google.com/site/nikobrummer/ (2010).

\bibitem{Caruana:2015:IMH:2783258.2788613}
R.~Caruana, et~al., Intelligible models for healthcare: Predicting pneumonia
  risk and hospital 30-day readmission, in: Proceedings of the 21th ACM SIGKDD
  International Conference on Knowledge Discovery and Data Mining, KDD '15,
  ACM, New York, NY, USA, 2015, pp. 1721--1730.
\newblock \href {http://dx.doi.org/10.1145/2783258.2788613}
  {\path{doi:10.1145/2783258.2788613}}.

\bibitem{zadrozny02}
B.~Zadrozny, et~al., Transforming classifier scores into accurate multiclass
  probability estimates, Proceeding of the Eight International Conference on
  Knowledge Discovery and Data Mining ({KDD}'02)\href
  {http://dx.doi.org/10.1145/775047.775151} {\path{doi:10.1145/775047.775151}}.

\bibitem{niculeskuMizil05predictingGoodProbabilities}
A.~Niculescu-Mizil, et~al., Predicting good probabilities with supervised
  learning, in: Proceedings of the 22nd International Conference on Machine
  Learning, Bonn, Germany, 2005, pp. 625--632.
\newblock \href {http://dx.doi.org/10.1145/1102351.1102430}
  {\path{doi:10.1145/1102351.1102430}}.

\bibitem{Gulcehre:2017:ILM:3103639.3103741}
C.~Gulcehre, et~al., On integrating a language model into neural machine
  translation, Comput. Speech Lang. 45~(C) (2017) 137--148.
\newblock \href {http://dx.doi.org/10.1016/j.csl.2017.01.014}
  {\path{doi:10.1016/j.csl.2017.01.014}}.

\bibitem{brummer06calibrationLanguage}
N.~Br\"ummer, et~al., On calibration of language recognition scores, in: Proc.
  of Odyssey, San Juan, Puerto Rico, 2006.

\bibitem{journals/corr/BojarskiTDFFGJM16}
M.~Bojarski, et~al., End to end learning for self-driving cars.

\bibitem{Lee2017TrainingCC}
K.~Lee, et~al., Training confidence-calibrated classifiers for detecting
  out-of-distribution samples, in: International Conference On Learning
  Representations, 2018.

\bibitem{deGroot83forecasters}
M.~H. deGroot, S.~E. Fienberg, The comparison and evaluation of forecasters,
  The Statistician 32 (1983) 12--22.

\bibitem{NIPS2017_7219}
B.~Lakshminarayanan, et~al., Simple and scalable predictive uncertainty
  estimation using deep ensembles, in: I.~Guyon, U.~V. Luxburg, S.~Bengio,
  H.~Wallach, R.~Fergus, S.~Vishwanathan, R.~Garnett (Eds.), Advances in Neural
  Information Processing Systems 30, Curran Associates, Inc., 2017, pp.
  6402--6413.

\bibitem{ramos18crossEntropy}
D.~Ramos, J.~Franco-Pedroso, A.~Lozano-Diez, J.~Gonzalez-Rodriguez,
  Deconstructing cross-entropy for probabilistic binary classifiers,
  Entropy~(3) (2018) 208.
\newblock \href {http://dx.doi.org/10.3390/e20030208}
  {\path{doi:10.3390/e20030208}}.

\bibitem{DBLP:journals/corr/GuoPSW17}
C.~Guo, et~al., On calibration of modern neural networks, in: D.~Precup, Y.~W.
  Teh (Eds.), Proceedings of the 34th International Conference on Machine
  Learning, Vol.~70 of Proceedings of Machine Learning Research, PMLR,
  International Convention Centre, Sydney, Australia, 2017, pp. 1321--1330.

\bibitem{DBLP:conf/icml/KuleshovFE18}
V.~Kuleshov, et~al., Accurate uncertainties for deep learning using calibrated
  regression, in: {ICML}, Vol.~80 of {JMLR} Workshop and Conference
  Proceedings, 2018, pp. 2801--2809.

\bibitem{pmlr-v80-kumar18a}
A.~Kumar, et~al., Trainable calibration measures for neural networks from
  kernel mean embeddings, in: J.~Dy, A.~Krause (Eds.), Proceedings of the 35th
  International Conference on Machine Learning, Vol.~80 of Proceedings of
  Machine Learning Research, PMLR, 2018, pp. 2805--2814.

\bibitem{1809.10877}
S.~{Seo}, et~al., Learning for single-shot confidence calibration in deep
  neural networks through stochastic inferences, in: 2019 IEEE/CVF Conference
  on Computer Vision and Pattern Recognition (CVPR), 2019, pp. 9022--9030.
\newblock \href {http://dx.doi.org/10.1109/CVPR.2019.00924}
  {\path{doi:10.1109/CVPR.2019.00924}}.

\bibitem{NIPS2017_7141}
A.~Kendall, et~al., What uncertainties do we need in bayesian deep learning for
  computer vision?, in: I.~Guyon, U.~V. Luxburg, S.~Bengio, H.~Wallach,
  R.~Fergus, S.~Vishwanathan, R.~Garnett (Eds.), Advances in Neural Information
  Processing Systems 30, Curran Associates, Inc., 2017, pp. 5574--5584.

\bibitem{fixing}
A.~Wu, et~al., Fixing variational bayes: Deterministic variational inference
  for bayesian neural networks, in: International Conference On Learning
  Representations, 2019.

\bibitem{Zadrozny:2001:OCP:645530.655658}
B.~Zadrozny, et~al., Obtaining calibrated probability estimates from decision
  trees and naive bayesian classifiers, in: Proceedings of the Eighteenth
  International Conference on Machine Learning, ICML '01, Morgan Kaufmann
  Publishers Inc., San Francisco, CA, USA, 2001, pp. 609--616.

\bibitem{Platt99probabilisticoutputs}
J.~C. Platt, Probabilistic outputs for support vector machines and comparisons
  to regularized likelihood methods, in: ADVANCES IN LARGE MARGIN CLASSIFIERS,
  MIT Press, 1999, pp. 61--74.

\bibitem{Naeini:2015:OWC:2888116.2888120}
M.~P. Naeini, et~al., Obtaining well calibrated probabilities using bayesian
  binning, in: Proceedings of the Twenty-Ninth AAAI Conference on Artificial
  Intelligence, AAAI'15, AAAI Press, 2015, pp. 2901--2907.

\bibitem{mcdropoutgal}
Y.~Gal, et~al., Dropout as a bayesian approximation: Representing model
  uncertainty in deep learning, in: Proceedings of the 33rd International
  Conference on International Conference on Machine Learning - Volume 48,
  ICML'16, JMLR.org, 2016, pp. 1050--1059.

\bibitem{Pereyra2017RegularizingNN}
G.~Pereyra, et~al., Regularizing neural networks by penalizing confident output
  distributions.

\bibitem{DBLP:journals/corr/abs-1805-05396}
T.~Chen, J.~Navratil, V.~Iyengar, K.~Shanmugam,
  \href{http://proceedings.mlr.press/v89/chen19c.html}{Confidence scoring using
  whitebox meta-models with linear classifier probes}, in: K.~Chaudhuri,
  M.~Sugiyama (Eds.), Proceedings of Machine Learning Research, Vol.~89 of
  Proceedings of Machine Learning Research, PMLR, 2019, pp. 1467--1475.
\newline\urlprefix\url{http://proceedings.mlr.press/v89/chen19c.html}

\bibitem{DeVries2018LearningCF}
T.~DeVries, et~al., Learning confidence for out-of-distribution detection in
  neural networks.

\bibitem{DBLP:journals/corr/GalG15a}
Y.~Gal, et~al., Bayesian convolutional neural networks with bernoulli
  approximate variational inference, in: International Conference On Learning
  Representations, Workshop track, 2016.

\bibitem{NIPS2015_5666}
D.~P. Kingma, et~al., Variational dropout and the local reparameterization
  trick, in: C.~Cortes, N.~D. Lawrence, D.~D. Lee, M.~Sugiyama, R.~Garnett
  (Eds.), Advances in Neural Information Processing Systems 28, Curran
  Associates, Inc., 2015, pp. 2575--2583.

\bibitem{1703.01961}
C.~Louizos, et~al., Multiplicative normalizing flows for variational {B}ayesian
  neural networks, in: D.~Precup, Y.~W. Teh (Eds.), Proceedings of the 34th
  International Conference on Machine Learning, Vol.~70 of Proceedings of
  Machine Learning Research, PMLR, 2017, pp. 2218--2227.

\bibitem{45819}
L.~Dinh, et~al., Density estimation using real nvp, in: International
  Conference on Learning Representations, 2017.

\bibitem{1505.05770}
D.~J. Rezende, et~al., Variational inference with normalizing flows, in:
  Proceedings of the 32Nd International Conference on Machine Learning - Volume
  37, ICML'15, JMLR.org, 2015, pp. 1530--1538.

\bibitem{Maaloe:2016:ADG:3045390.3045543}
L.~Maal{\o}e, et~al., Auxiliary deep generative models, in: Proceedings of the
  33rd International Conference on International Conference on Machine Learning
  - Volume 48, ICML'16, JMLR.org, 2016, pp. 1445--1454.

\bibitem{DBLP:journals/corr/abs-1805-10377}
Y.~Zhang, et~al., Variational measure preserving flows, CoRR abs/1805.10377.
\newblock \href {http://arxiv.org/abs/1805.10377} {\path{arXiv:1805.10377}}.

\bibitem{1206.1901}
R.~M. Neal, {MCMC} using {Hamiltonian} dynamics, Handbook of Markov Chain Monte
  Carlo 54 (2010) 113--162.

\bibitem{Bishop:2006:PRM:1162264}
C.~M. Bishop, Pattern Recognition and Machine Learning (Information Science and
  Statistics), Springer-Verlag, Berlin, Heidelberg, 2006.

\bibitem{Gal2016Uncertainty}
Y.~Gal, Uncertainty in deep learning, Ph.D. thesis, University of Cambridge
  (2016).

\bibitem{NIPS2005_2857}
E.~Snelson, et~al., Sparse gaussian processes using pseudo-inputs, in:
  Y.~Weiss, B.~Sch\"{o}lkopf, J.~C. Platt (Eds.), Advances in Neural
  Information Processing Systems 18, MIT Press, 2006, pp. 1257--1264.

\bibitem{NIPS2018_7979}
M.~Havasi, et~al., Inference in deep gaussian processes using stochastic
  gradient hamiltonian monte carlo, in: S.~Bengio, H.~Wallach, H.~Larochelle,
  K.~Grauman, N.~Cesa-Bianchi, R.~Garnett (Eds.), Advances in Neural
  Information Processing Systems 31, Curran Associates, Inc., 2018, pp.
  7506--7516.

\bibitem{Chen:2014:SGH:3044805.3045080}
T.~Chen, et~al., Stochastic gradient hamiltonian monte carlo, in: Proceedings
  of the 31st International Conference on International Conference on Machine
  Learning - Volume 32, ICML'14, JMLR.org, 2014, pp. II--1683--II--1691.

\bibitem{betancourt2017conceptual}
M.~Betancourt, A conceptual introduction to hamiltonian monte carlo,
  arxiv:1701.02434 (2017).

\bibitem{1312.6114}
D.~P. Kingma, et~al., Auto-encoding variational bayes, in: International
  Conference on Learning Representations, 2014.

\bibitem{1401.4082}
D.~J. Rezende, et~al., Stochastic backpropagation and approximate inference in
  deep generative models, in: E.~P. Xing, T.~Jebara (Eds.), Proceedings of the
  31st International Conference on Machine Learning, Vol.~32 of Proceedings of
  Machine Learning Research, PMLR, Bejing, China, 2014, pp. 1278--1286.

\bibitem{WahCUB_200_2011}
C.~Wah, et~al., {The Caltech-UCSD Birds-200-2011 Dataset}, Tech. Rep.
  CNS-TR-2011-001, California Institute of Technology (2011).

\bibitem{KrauseStarkDengFei-Fei_3DRR2013}
J.~Krause, M.~Stark, J.~Deng, L.~Fei-Fei, 3d object representations for
  fine-grained categorization, in: 4th International IEEE Workshop on 3D
  Representation and Recognition (3dRR-13), Sydney, Australia, 2013.

\bibitem{cifar100}
A.~Krizhevsky, et~al.,
  \href{http://www.cs.toronto.edu/~kriz/cifar.html}{Cifar-100 (canadian
  institute for advanced research)}.
\newline\urlprefix\url{http://www.cs.toronto.edu/~kriz/cifar.html}

\bibitem{cifar10}
A.~Krizhevsky, et~al.,
  \href{http://www.cs.toronto.edu/~kriz/cifar.html}{Cifar-10 (canadian
  institute for advanced research)}.
\newline\urlprefix\url{http://www.cs.toronto.edu/~kriz/cifar.html}

\bibitem{noauthororeditor}
Y.~o. Netzer, Reading digits in natural images with unsupervised feature
  learning.

\bibitem{Cao18}
Q.~Cao, others., Vggface2: A dataset for recognising faces across pose and age,
  in: International Conference on Automatic Face and Gesture Recognition, 2018.

\bibitem{Eidinger:2014:AGE:2771306.2772049}
E.~Eidinger, et~al., Age and gender estimation of unfiltered faces, Trans.
  Info. For. Sec. 9~(12) (2014) 2170--2179.
\newblock \href {http://dx.doi.org/10.1109/TIFS.2014.2359646}
  {\path{doi:10.1109/TIFS.2014.2359646}}.

\bibitem{vgg_1409.1556}
K.~Simonyan, et~al., Very deep convolutional networks for large-scale image
  recognition, in: International Conference On Learning Representations, 2015.

\bibitem{DBLP:journals/corr/HeZRS15}
K.~He, et~al., Deep residual learning for image recognition, in: 2016 {IEEE}
  Conference on Computer Vision and Pattern Recognition, 2016, pp. 770--778.
\newblock \href {http://dx.doi.org/10.1109/CVPR.2016.90}
  {\path{doi:10.1109/CVPR.2016.90}}.

\bibitem{1603.05027}
K.~He, et~al., Identity mappings in deep residual networks, in: ECCV, 2016.

\bibitem{dpn_1707.01629}
Y.~Chen, et~al., Dual path networks, in: I.~Guyon, U.~V. Luxburg, S.~Bengio,
  H.~Wallach, R.~Fergus, S.~Vishwanathan, R.~Garnett (Eds.), Advances in Neural
  Information Processing Systems 30, Curran Associates, Inc., 2017, pp.
  4467--4475.

\bibitem{resnext_1611.05431}
S.~Xie, et~al., Aggregated residual transformations for deep neural networks,
  2017 IEEE Conference on Computer Vision and Pattern Recognition (CVPR) (2017)
  5987--5995.

\bibitem{Sandler_2018_CVPR}
M.~Sandler, et~al., Mobilenetv2: Inverted residuals and linear bottlenecks, in:
  The IEEE Conference on Computer Vision and Pattern Recognition (CVPR), 2018.

\bibitem{Hu18}
J.~Hu, et~al., Squeeze-and-excitation networks, 2018 IEEE/CVF Conference on
  Computer Vision and Pattern Recognition (2018) 7132--7141.

\bibitem{adam_1412.6980}
D.~P. Kingma, J.~Ba, Adam: A method for stochastic optimization (2014).

\bibitem{43405}
I.~Goodfellow, et~al., Explaining and harnessing adversarial examples, in:
  International Conference on Learning Representations, 2015.

\bibitem{NIPS2017_6949}
Y.~Gal, et~al.,
  \href{http://papers.nips.cc/paper/6949-concrete-dropout.pdf}{Concrete
  dropout}, in: I.~Guyon, U.~V. Luxburg, S.~Bengio, H.~Wallach, R.~Fergus,
  S.~Vishwanathan, R.~Garnett (Eds.), Advances in Neural Information Processing
  Systems 30, Curran Associates, Inc., 2017, pp. 3581--3590.
\newline\urlprefix\url{http://papers.nips.cc/paper/6949-concrete-dropout.pdf}

\bibitem{1805.10522}
G.-L. Tran, et~al., Calibrating deep convolutional gaussian processes, in:
  K.~Chaudhuri, M.~Sugiyama (Eds.), Proceedings of Machine Learning Research,
  Vol.~89 of Proceedings of Machine Learning Research, PMLR, 2019, pp.
  1554--1563.

\bibitem{1805.10915}
D.~Milios, othes, Dirichlet-based gaussian processes for large-scale calibrated
  classification, in: S.~Bengio, H.~Wallach, H.~Larochelle, K.~Grauman,
  N.~Cesa-Bianchi, R.~Garnett (Eds.), Advances in Neural Information Processing
  Systems 31, Curran Associates, Inc., 2018, pp. 6005--6015.

\bibitem{NIPS2016_6581}
D.~P. Kingma, et~al., Improved variational inference with inverse
  autoregressive flow, in: D.~D. Lee, M.~Sugiyama, U.~V. Luxburg, I.~Guyon,
  R.~Garnett (Eds.), Advances in Neural Information Processing Systems 29,
  Curran Associates, Inc., 2016, pp. 4743--4751.

\bibitem{Huang2018NeuralAF}
C.-W. Huang, et~al., Neural autoregressive flows, in: ICML, 2018.

\bibitem{Berg2018SylvesterNF}
{Van Den Berg}, et~al., Sylvester normalizing flows for variational inference,
  in: A.~Globerson, A.~Globerson, R.~Silva (Eds.), 34th Conference on
  Uncertainty in Artificial Intelligence 2018, UAI 2018, 34th Conference on
  Uncertainty in Artificial Intelligence 2018, UAI 2018, Association For
  Uncertainty in Artificial Intelligence (AUAI), 2018, pp. 393--402.

\bibitem{DBLP:conf/iconip/AgakovB04a}
F.~V. Agakov, et~al., An auxiliary variational method, in: Neural Information
  Processing, 11th International Conference, {ICONIP} 2004, Calcutta, India,
  November 22-25, 2004, Proceedings, 2004, pp. 561--566.

\bibitem{1511.02386}
R.~Ranganath, et~al., Hierarchical variational models, in: M.~F. Balcan, K.~Q.
  Weinberger (Eds.), Proceedings of The 33rd International Conference on
  Machine Learning, Vol.~48 of Proceedings of Machine Learning Research, PMLR,
  New York, New York, USA, 2016, pp. 324--333.

\bibitem{pmlr-v80-cremer18a}
C.~Cremer, et~al., Inference suboptimality in variational autoencoders, in:
  J.~Dy, A.~Krause (Eds.), Proceedings of the 35th International Conference on
  Machine Learning, Vol.~80 of Proceedings of Machine Learning Research, PMLR,
  Stockholmsmässan, Stockholm Sweden, 2018, pp. 1086--1094.

\bibitem{DBLP:journals/corr/abs-1805-08913}
R.~Shu, et~al., Amortized inference regularization, in: S.~Bengio, H.~Wallach,
  H.~Larochelle, K.~Grauman, N.~Cesa-Bianchi, R.~Garnett (Eds.), Advances in
  Neural Information Processing Systems 31, Curran Associates, Inc., 2018, pp.
  4393--4402.

\bibitem{pmlr-v80-kim18e}
Y.~Kim, et~al., Semi-amortized variational autoencoders, in: J.~Dy, A.~Krause
  (Eds.), Proceedings of the 35th International Conference on Machine Learning,
  Vol.~80 of Proceedings of Machine Learning Research, PMLR, Stockholmsmässan,
  Stockholm Sweden, 2018, pp. 2683--2692.

\bibitem{knoblauch2019generalized}
J.~Knoblauch, J.~Jewson, T.~Damoulas, Generalized variational inference: Three
  arguments for deriving new posteriors (2019).
\newblock \href {http://arxiv.org/abs/1904.02063} {\path{arXiv:1904.02063}}.

\end{thebibliography}

\end{document}